\documentclass[sigconf]{acmart}
\usepackage{microtype}
\usepackage{graphicx}
\usepackage{subfigure}
\usepackage{booktabs} 
\usepackage{url}

\usepackage{hyperref}
\usepackage{utfsym}
\usepackage{algorithm}
\usepackage{algorithmic}

\usepackage{amsmath}
\usepackage{amsthm}
\usepackage[capitalize,noabbrev]{cleveref}

\theoremstyle{plain}
\newtheorem{theorem}{Theorem}

\newtheorem{lemma}{Lemma}

\theoremstyle{definition}
\newtheorem{definition}{Definition}
\newtheorem{assumption}{Assumption}
\newtheorem{property}{Property}
\theoremstyle{remark}

\newcommand{\tabincell}[2]{\begin{tabular}{@{}#1@{}}#2\end{tabular}}
\usepackage{balance}
\usepackage{footmisc}

\AtBeginDocument{%
  \providecommand\BibTeX{{%
    \normalfont B\kern-0.5em{\scshape i\kern-0.25em b}\kern-0.8em\TeX}}}

\copyrightyear{2024}
\acmYear{2024}
\setcopyright{rightsretained}
\acmConference[KDD '24]{Proceedings of the 30th ACM SIGKDD Conference on Knowledge Discovery and Data Mining}{August 25--29, 2024}{Barcelona, Spain} \acmBooktitle{Proceedings of the 30th ACM SIGKDD Conference on Knowledge Discovery and Data Mining (KDD '24), August 25--29, 2024, Barcelona, Spain}\acmDOI{10.1145/3637528.3671721}
\acmISBN{979-8-4007-0490-1/24/08}

\makeatletter
\gdef\@copyrightpermission{
  \begin{minipage}{0.3\columnwidth}
   \href{https://creativecommons.org/licenses/by-nc-sa/4.0/}{\includegraphics[width=0.90\textwidth]{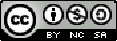}}
  \end{minipage}\hfill
  \begin{minipage}{0.7\columnwidth}
   \href{https://creativecommons.org/licenses/by-nc-sa/4.0/}{This work is licensed under a Creative Commons Attribution-NonCommercial-ShareAlike International 4.0 License.}
  \end{minipage}
  \vspace{5pt}
}
\makeatother
\begin{document}

\title{POND: Multi-Source Time Series Domain Adaptation with Information-Aware Prompt Tuning}

\author{Junxiang Wang}
\authornote{Both authors contributed equally to this research.}
\affiliation{%
  \institution{NEC Labs America}
  \streetaddress{4 Independence Way}
  \city{Princeton}
  \state{New Jersey}
  \country{USA}
  \postcode{08540}
}
\author{Guangji Bai}
\authornotemark[1]
\affiliation{%
  \institution{Emory University}
  \city{Atlanta}
  \state{Georgia}
  \country{USA}
  \postcode{30322}
}
\author{Wei Cheng}
\affiliation{%
\institution{NEC Labs America}
  \streetaddress{4 Independence Way}
  \city{Princeton}
  \state{New Jersey}
  \country{USA}
  \postcode{08540}
  }
\author{Zhengzhang Chen}
\affiliation{%
\institution{NEC Labs America}
  \streetaddress{4 Independence Way}
  \city{Princeton}
  \state{New Jersey}
  \country{USA}
  \postcode{08540}
  }
\author{Liang Zhao}
\affiliation{%
  \institution{Emory University}
  \city{Atlanta}
  \state{Georgia}
  \country{USA}
  \postcode{30322}
}
\author{Haifeng Chen}
\affiliation{%
\institution{NEC Labs America}
  \streetaddress{4 Independence Way}
  \city{Princeton}
  \state{New Jersey}
  \country{USA}
  \postcode{08540}
  }
\renewcommand{\shortauthors}{Junxiang Wang et al.}







\begin{abstract}
Time series domain adaptation stands as a pivotal and intricate challenge with diverse applications, including but not limited to human activity recognition, sleep stage classification, and machine fault diagnosis. Despite the numerous domain adaptation techniques proposed to tackle this complex problem, they primarily focus on domain adaptation from a single source domain. Yet, it is more crucial to investigate domain adaptation from multiple domains due to the potential for greater improvements. To address this, three important challenges need to be overcome: 1). The lack of exploration to utilize domain-specific information for domain adaptation, 2). The difficulty to learn domain-specific information that changes over time, and 3). The difficulty to evaluate learned domain-specific information. In order to tackle these challenges simultaneously, in this paper, we introduce PrOmpt-based domaiN Discrimination (POND), the first framework to utilize prompts for time series domain adaptation. Specifically, to address Challenge 1, we extend the idea of prompt tuning to time series analysis and learn prompts to capture common and domain-specific information from all source domains. To handle Challenge 2, we introduce a conditional module for each source domain to generate prompts from time series input data. For Challenge 3, we propose two criteria to select good prompts, which are used to choose the most suitable source domain for domain adaptation. The efficacy and robustness of our proposed POND model are extensively validated through experiments across 50 scenarios encompassing four datasets. Experimental results demonstrate that our proposed POND model outperforms all state-of-the-art comparison methods by up to $66\%$ on the F1-score.
\end{abstract}
\maketitle
\section{Introduction}
\label{sec: introduction}
Due to the prevalence of time series sensor data, time series analysis has found applications in various real-world scenarios, including human activity recognition \cite{anguita2013public}, sleep stage classification \cite{zhao2017learning}, and machine fault diagnosis~\cite{wang2023incremental,wang2023interdependent,lessmeier2016condition}. In these applications, time series data are measured under different subjects, operating conditions, or sensor configurations (\textit{i.e.}, domains). In other words, time series analysis should be conducted across different domains. Unfortunately, the labels of time series data are difficult to collect due to the expensive costs of the labeling process \cite{wilson2020multi}. To mitigate labeling costs, researchers aim to leverage labeled data from some domains (\textit{i.e.}, source domains) to infer labels for unlabeled data in other domains (\textit{i.e.}, target domains) \cite{wang2018multi}, which is defined as a time series domain adaptation problem. For example, the goal of the transponder fault diagnosis problem is to detect the working statuses of transponders (\textit{i.e.}, normal or abnormal) based on fiber-optic signals. In this problem, the model is trained under certain working modes (\textit{e.g.}, single mode) using labeled time series data, and then this trained model is applied to other working modes (\textit{e.g.}, multimode).\\
 \indent However, the time series domain adaptation problem is highly challenging due to complex dynamic time series patterns, distribution shift (\textit{i.e.}, different distributions of inputs among different domains), and possible label shift (\textit{i.e.}, different distributions of labels among different domains) \cite{bai2022temporal,he2023domain,cai2024continuous}. These challenges have been extensively investigated by researchers, leading to the proposal of various methods to address the domain gap, such as kernel matching \cite{liu2021adversarial}, context information alignment \cite{lai2023context}, and temporal-spectral fusion \cite{yang2022unsupervised}. Most existing methods, however, primarily focus on domain adaptation from a single source domain. Yet, it is more crucial to investigate it from multiple sources. This is because the more source domains are utilized, the greater potential improvements it can achieve. For instance, the collection of labeled signal data from more modes facilitates a better understanding of transponder statuses. Despite the importance of the multi-source domain adaptation problem, it is rarely explored in previous literature and requires attention and extensive investigations from researchers.\\
\indent In order to effectively handle the multi-source time series domain adaptation problem, three important challenges need to be overcome: \textbf{1. The lack of exploration to utilize domain-specific information for domain adaptation.} Existing domain adaptation methods primarily focus on learning a common feature extractor to encode time series inputs from different source domains into domain-invariant representations, and then apply this feature extractor to the target domain \cite{peng2019domain,li2022towards,luo2018tinet,jin2022domain,wilson2020multi}. While this strategy has its rationale, it often overlooks domain-specific information (\textit{i.e.}, information unique to a specific time series domain), such as global trends, local trends, and temporal patterns. Such domain-specific information is valuable to evaluate which source domains are more suitable for adaptation to the target domain. \textbf{2. The difficulty to learn domain-specific information that changes over time.} While it is important to capture domain-specific information for better domain adaptation, such information can be dynamically changing, which is extremely difficult to capture. In the example of the transponder fault diagnosis problem, different domains generate different distributions of fiber-optic signals, which are important domain-specific information to capture. However, such distributions can be shifted drastically when the transponder suddenly suffers from a failure. \textbf{3. The difficulty to evaluate learned domain-specific information.} Not only is learning domain-specific information difficult, but it is also challenging to evaluate learned domain-specific information. In other words, it is unclear whether learned domain-specific information accurately reflects the true one. This ambiguity arises because domain-specific information is often associated with unique but inexplicable underlying patterns. Unlike images and languages with human-recognizable features, such time series patterns are difficult for humans to understand \cite{luo2023time}. Consequently, it becomes challenging, if not impossible, for humans to evaluate whether learned domain-specific information matches such time series patterns.\\
 \begin{figure}
    \centering
\includegraphics[width=\linewidth]{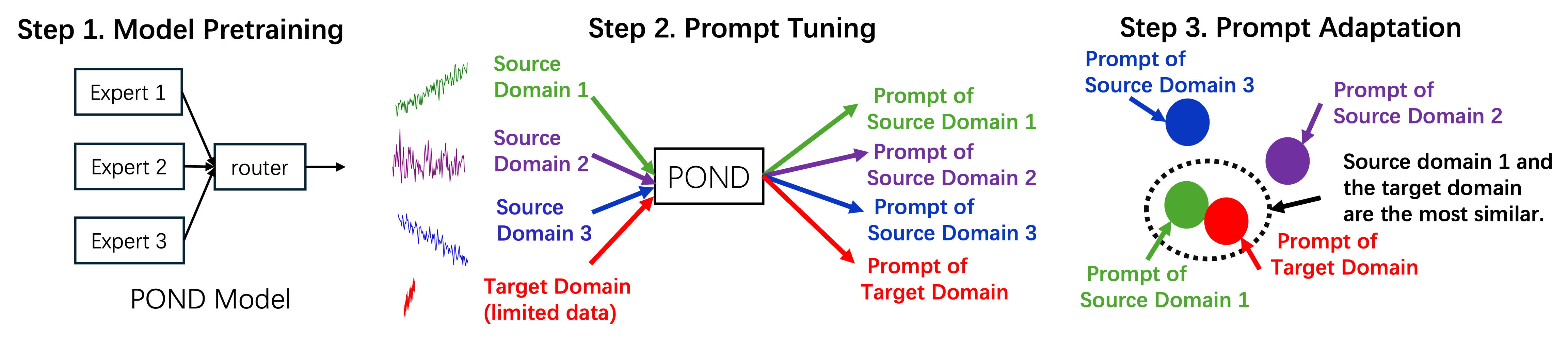}
    \caption{Pipeline of our proposed POND model: Step 1 pretrains the proposed POND model; Step 2 learns prompts of all source domains and the target domain; Step 3 utilizes learned prompts to select the most similar source domain to the target domain for domain adaptation.}
    \label{fig:pipeline}
    \vspace{-0.7cm}
\end{figure}
 \indent In order to tackle these three challenges simultaneously, we propose PrOmpt-based domaiN Discrimination (POND), the first framework to utilize prompts for time series domain adaptation to our knowledge. Its pipeline is shown in Figure \ref{fig:pipeline}, which consists of three steps: model pertaining, prompt tuning, and prompt adaptation. Specifically, to address Challenge 1, we extend the idea of prompt tuning to time series analysis and learn prompts to capture common and domain-specific information. To handle Challenge 2, we introduce a conditional module for each source domain to generate prompts from time series input data. For Challenge 3, we propose two criteria to choose good prompts, which are used to select the most suitable source domain for domain adaptation (\textit{i.e.}, prompt adaptation). Our contributions can be summarized as follows:
\begin{itemize}
\item \textbf{Propose a flexible prompt generator to learn domain-specific information.} We extend the idea of prompt tuning to time series analysis to capture information specific to source domains. However, traditional prompts have limited flexibility in learning domain-specific information that evolves over time. To address this limitation, we introduce a conditional module that generates prompts parameterized by a neural network to capture domain-specific information. Theoretical analysis also demonstrates the superiority of our proposed prompt generator over traditional prompt tuning.

\item \textbf{Develop two criteria for selecting good prompts.} We propose two criteria, fidelity and distinction, to ensure that prompts accurately capture domain-specific information from all source domains. Fidelity is achieved by maximizing the mutual information between prompts and labels, while distinction is achieved by minimizing the mutual information between prompts from different source domains. Theoretical guarantees establish that our generated prompts maintain fidelity and introduce new information.

\item \textbf{Present an efficient algorithm with a robust architecture.} We introduce a simple yet effective optimization algorithm based on meta-learning to efficiently learn the objective. Additionally, we leverage the Mixture of Experts (MoE) technique to enhance the robustness of our proposed POND model.

\item \textbf{Conduct comprehensive experiments on multiple benchmark datasets.} Extensive experiments across 50 scenarios on four benchmark datasets demonstrate the effectiveness and robustness of our proposed POND model. Experimental results indicate that our proposed POND model outperforms all state-of-the-art comparison methods by up to $66\%$ on the F1-score.
\end{itemize}
\section{Related Work}
Previous research related to this study can be categorized into two main areas: time series domain adaptation and Large Language Models (LLMs) for time series.\\
\indent\textbf{Time Series Domain Adaptation:} Works in this domain can be classified into Unsupervised Domain Adaptation (UDA) and supervised methods.\\
\indent UDA is a common approach, particularly beneficial as it does not rely on labels in the target domain. For example, Liu and Xue introduced the Adversarial Spectral Kernel Matching (AdvSKM) approach, employing a specialized hybrid spectral kernel network to redefine the Maximum Mean Discrepancy (MMD) metric \cite{liu2021adversarial}. Lai et al. aligned context information between different time series domains using a Markov decision process formulation and employed deep reinforcement learning for anomaly detection \cite{lai2023context}. He et al. addressed feature and label shifts between the source and target domains using temporal and frequency features \cite{he2023domain}. Other notable approaches include autoregressive models \cite{ragab2022self}, sparse associative structure alignment \cite{cai2021time}, variational methods \cite{li2022towards, purushotham2016variational}, contrastive learning \cite{yue2022ts2vec}, and temporal-spectral fusion \cite{yang2022unsupervised}.\\
\indent In addition to UDA, other methods transfer time series knowledge in a supervised manner. For instance, Jin et al. proposed an attention-based shared module to learn common latent features, incorporating a domain discriminator retaining domain-specific features across multiple domains \cite{jin2022domain}. Wilson et al. leveraged target-domain label distributions to enhance model performance with benefits from multi-source time series data \cite{wilson2020multi}. However, to our knowledge, all existing time series domain adaptation methods neglect domain-specific information such as unique temporal patterns, which could potentially be utilized for better domain adaptation.\\
\indent \textbf{LLMs for Time Series:} Large Language Models (LLMs) have shown excellent performance in various Natural Language Processing (NLP) tasks such as natural language inference, question answering, and named entity recognition \cite{zhao2023survey}. Recent research has extended LLMs to address time series problems, generally falling into two classes: prompt tuning and fine-tuning.\\
\indent In prompt tuning methods, pretrained LLMs use prompts (\textit{i.e.}, a sequence of tokens prepended to the time series input) to learn specific downstream tasks. For example, Xue and Salim proposed PromptCast, a novel approach that transforms numerical input and output into prompts and frames the time series forecasting task in a sentence-to-sentence manner \cite{xue2022promptcast}. Cao et al. presented the TEMPO framework, which decomposed complex interactions between trend, seasonal, and residual components, introducing selection-based prompts to facilitate distribution adaptation in non-stationary time series \cite{cao2023tempo}. Jin et al. proposed the TIME-LLM framework, reprogramming the input time series with text prototypes before feeding it into a frozen LLM to align the two modalities, with Prompt-as-Prefix (PaP) introduced to enrich the input context and guide the transformation of the reprogrammed input \cite{jin2023time}. LLMTime highlighted the efficacy of LLMs as zero-shot learners by encoding numbers into texts as prompts and sampling possible extrapolations as prompt completions \cite{gruver2023large}. Sun et al. proposed the TEST model, training an encoder to embed time series tokens with contrastive learning and aligning text prototypes with time series, utilizing prompts to adapt LLMs to different time series tasks \cite{sun2023test}.\\
\indent In contrast, fine-tuning is the other type of method to adapt LLMs to time series, adjusting some components while keeping others frozen. For example, Zhou et al. presented the OFA framework, where only the embedding and normalization layers of LLMs were fine-tuned, while self-attention and feed-forward layers remained frozen \cite{zhou2023one}. Chang et al. proposed the Llm4ts framework, fine-tuning in two stages: first, supervised fine-tuning to orient the LLM towards time series data, followed by task-specific downstream fine-tuning \cite{chang2023llm4ts}. For more information, please refer to the recent survey paper by Jin et al. \cite{jin2023lm4ts}. While these methods transfer knowledge from LLMs to the time series domain, they do not address the time series domain adaptation problem, where knowledge from the source time series domain, rather than text, is transferred to the target domain.
\vspace{-0.3cm}
\section{Problem Setup}
\begin{table}[h]
\small
    \caption{Important notations and Descriptions.}
    \centering
    \begin{tabular}{c|c}
    \hline
    Notations & Descriptions\\\hline
         $S_i$ & The $i$-th source domain \\
         $T$ & Target domain \\
         $C$ & Class set \\ 
         $(X^{(S_i)}_j,Y^{(S_i)}_j)$ & The $j$-th time series pair for $S_i$\\
         $(X^{(T)}_j,Y^{(T)}_j)$ & The $j$-th time series pair for  $T$\\
         $Y^{(S_i)}, Y^{(T)}$ & Label sets for $S_i$ and $T$\\
         $P$ & Common prompt\\
         $\Delta P^{(S_i)}$ & Domain-level prompt for $S_i$\\
         $\Delta P^{(S_i)}_j$ & Instance-level prompt generated by  $X^{(S_i)}_j$ for $S_i$\\ \hline
    \end{tabular}
    \label{tab:notation}
    \vspace{-0.3cm}
\end{table}
 \indent In this section, we mathematically formulate the multi-source time series domain adaptation problem. Important notations are shown in Table \ref{tab:notation}. Given $M$ source time series domains $S_i (i=1,\cdots,M)$ and a target domain $T$, their $j$-th time series inputs are denoted as $X^{(S_i)}_j \sim p(X|Y^{(S_i)}_j)$ and $X^{(T)}_j \sim p(X|Y^{(T)}_j)$, respectively, where $Y^{(S_i)}_j$ and $Y^{(T)}_j$ are corresponding labels of $X^{(S_i)}_j$ and $X^{(T)}_j$, respectively. Here, $X^{(S_i)}_j, X^{(T)}_j \in \mathbb{R}^{n \times L}$, where $n$ is the number of channels and $L$ is the sequence length. The labels $Y^{(S_i)}_j, Y^{(T)}_j \in C = \{c_1, c_2, \cdots, c_K\}$, where $c_i (i=1,\cdots, |C|)$ represents a label class, and the number of classes is $|C|$. $Y^{(S_i)} = \{Y^{(S_i)}_j\}$ and $Y^{(T)} = \{Y^{(T)}_j\}$ are the label sets for the source domain $S_i$ and the target domain $T$, respectively. Sets $X^{(S_i)} = \{X^{(S_i)}_j\}$ and $X^{(T)} = \{X^{(T)}_j\}$ represent the input sets for the source domain $S_i$ and the target domain $T$, respectively. We assume that the labeled time series of all source domains $S_i (i=1,\cdots, M)$ are abundant, but the labeled time series are limited in the target domain $T$. Then the multi-source time series domain adaptation problem is formulated as follows:\\
\indent \textbf{Problem Formulation:} Given the time series input sets $X^{(S_i)}$ and label sets $Y^{(S_i)} (i=1,2,\dots, M)$ of $M$ source domains, and the time series input set $X^{(T)}$ of the target domain $T$, the goal of the problem is to predict the label set $Y^{(T)}$ by learning the mapping $F$:
\begin{align*}
    F: X_i^{(T)} \rightarrow Y_i^{(T)}
\end{align*}
\indent Our problem formulation is very flexible: the time series input can be either univariate (\textit{i.e.}, $N=1$) or multivariate (\textit{i.e.}, $N>1$); the time series domain adaptation can be from a single source (\textit{i.e.}, $M=1$) or multiple sources (\textit{i.e.}, $M>1$); the classification problem can be either binary (\textit{i.e.}, $K=2$) or multi-class (\textit{i.e.}, $K>2$).
\section{Prompt-based Domain Discrimination}
In this section, we present our POND model to address the multi-source time series domain adaptation problem. 
\subsection{The Flexible Prompt Generator}
\label{sec:prompt generator}

The goal of this section is to explore methods for learning information that changes over time from different source domains for domain adaptation (\textit{i.e.}, tackling Challenges 1 and 2). Most existing papers propose various strategies to extract domain-invariant representations from all source domains by making different domains indistinguishable \cite{peng2019domain,li2022towards,jin2022domain,wilson2020multi,zhao2020multi}. However, this idea may discard domain-specific information from multiple source domains, which indicates which source domain is most similar to the target domain. To address this, a natural solution is to directly learn domain-specific information from the labeled time series pair $(X^{(S_i)}_j, Y^{(S_i)}_j)$. This motivates us to utilize prompt tuning to learn domain-specific information, which was first introduced by the NLP community and demonstrated impressive success in many NLP tasks \cite{brown2020language,ling2023domain,lester2021power}. Compared with other domain adaptation techniques, prompt tuning has three advantages: firstly, prompts are adjusted via gradients by labeled data from multiple source domains, which offer domain-specific information; secondly, prompt tuning leverages small amounts of labeled data effectively for adaptation, which is suitable for the target domain with limited labeled data \cite{lester2021power}; thirdly, prompts can be utilized as a heuristic to select the most similar source domain to the target domain for adaptation.\\
The prompt, which is extended from NLP to time series, is defined as a learnable vector that prepends to the time series input to learn domain-specific information by the labeled pair $(X^{(S_i)}_j, Y^{(S_i)}_j)$. Mathematically, let $P^{(S_i)} \in \mathbb{R}^{n \times m}$ be the prompt of the source domain $S_i$, where $m$ is the prompt length. Then, for the $j$-th time series input $X^{(S_i)}_j$, any time series model takes $[P^{(S_i)}, X^{(S_i)}_j]$ (\textit{i.e.}, the concatenation of $P^{(S_i)}$ and $X^{(S_i)}_j$) as its model input. We decompose $P^{(S_i)}$ into two components:
\begin{align*}
    P^{(S_i)} = P + \Delta P^{(S_i)}
\end{align*}
where $P \in \mathbb{R}^{n \times m}$ is a common prompt to learn the common characteristics of all source domains, which can also be directly applied to the target domain $T$, and $\Delta P^{(S_i)} \in \mathbb{R}^{n \times m}$ is a prompt to learn domain-specific information (\textit{i.e.}, information unique to the source domain $S_i$), which will be utilized to select the most similar source domain to the target domain $T$.

While the domain-specific prompt $\Delta P^{(S_i)}$ is potentially effective to learn domain-specific information about the source domain $S_i$ (\textit{i.e.}, address Challenge 1), it cannot directly address Challenge 2. This is because $\Delta P^{(S_i)}$ is time-independent and has little freedom to capture time-dependent domain-specific information (\textit{e.g.}, distribution shifts of fiber-optic signals). To tackle this, instead of using a fixed prompt, we learn such domain-specific information by prompts generated from the time series input. This is because the time series input usually contains rich time-dependent information (\textit{e.g.}, time series distributions and trends). Specifically, we introduce a conditional module $g^{(S_i)}$, parameterized by a neural network, to generate instance-level prompts based on time series instances:
\begin{align*}
    \Delta P^{(S_i)}_{j} = g^{(S_i)}(X^{(S_i)}_{j}; \zeta) \in \mathbb{R}^{m \times n}
\end{align*}
where $\Delta P^{(S_i)}_{j}$ is the instance-level prompt generated by the time series input $X^{(S_i)}_{j}$ and a random variable $\zeta$, and the domain-level prompt $\Delta P^{(S_i)}$ is the aggregation of all instance-level prompts $\Delta P^{(S_i)}_j$ (\textit{e.g.}, $\Delta P^{(S_i)} = \frac{1}{|S_i|}\sum_{j=1}^{|S_i|} \Delta P^{(S_i)}_j$). For any time series input $X^{(S_i)}_{j}$, its corresponding prompt is formulated as $P + \Delta P^{(S_i)}_{j}$.

Our proposed prompt generator $g^{(S_i)}$ conditionally generates instance-level prompts for specific time series inputs, which intuitively has more freedom of expression to learn domain-specific information than the traditional prompt tuning. More theoretical investigations are provided to illustrate the power of the common prompt $P$ and the prompt generator $g^{(S_i)}$ in Section \ref{sec: discussion}.
\begin{figure}
    \centering
    \includegraphics[width=0.7\linewidth]{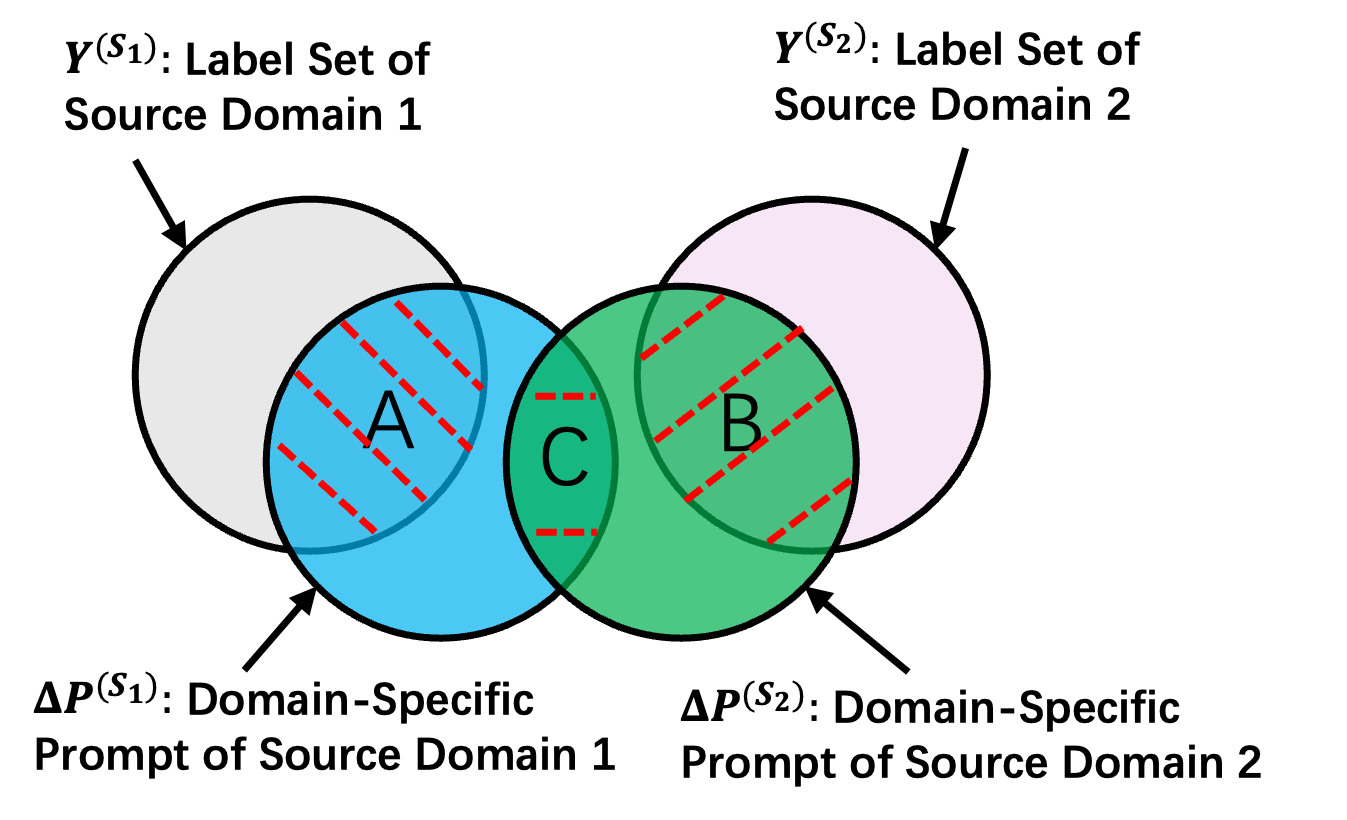}
    \vspace{-0.3cm}
    \caption{Illustration of two criteria: high fidelity and high distinction. Fidelity and distinction are represented as areas of $A+B$ and $C$, respectively.}
    \label{fig:criteria}
    \vspace{-0.5cm}
\end{figure}
\subsection{Two Important Criteria for Good Prompts}
\label{sec: criteria}
In the previous section, we extended prompt tuning to capture information on specific time series domains. While prompts are easy to recognize in computer vision and natural language fields, the learned prompts of time series data are not recognizable to humans, making it hard, if not impossible, to evaluate whether prompts are good enough to learn information for time series data. For example, a hard prompt consists of natural language that clearly describes the task at hand, explicitly asks the model for some result or action, and makes it easy to understand why the prompt elicited such behavior from the model \cite{lester2021power}. In contrast, the learned prompts of specific time series domains are visualized as extra time segments, which are difficult to understand by humans. Moreover, there is a lack of exploration on what constitutes a good prompt that captures domain-specific information without human-engineering priors. From our perspective, ideal prompts to capture domain-specific information should maintain high fidelity and high distinction, as illustrated in Figure \ref{fig:criteria}: high fidelity suggests large overlaps between the learned domain-specific prompts and label information (\textit{i.e.}, large $A+B$ in Figure \ref{fig:criteria}), and high distinction implies small overlaps among domain-specific prompts of different source domains (\textit{i.e.}, small $C$ in Figure \ref{fig:criteria}). They are introduced in details as follows:\\
\indent \textbf{High Fidelity.} One important criterion for the prompt generator $g^{(S_i)}$ is fidelity (\textit{i.e.}, the generated prompt $\Delta P^{(S_i)}_{j}$ preserves the domain-specific information of the source domain $S_i$). Motivated by the theory of information bottleneck \cite{tishby2000information}, high fidelity is defined as the large mutual information between $\Delta P^{(S_i)}_{j}$ and $Y^{(S_i)}_{j}$, which should be maximized:
\begin{align}
\max\sum_{i=1}^M\sum_{j=1}^{\vert S_i\vert} MI(\Delta P^{(S_i)}_j,Y^{(S_i)}_j),
\label{eq: maximize mutual information}
\end{align}
where $MI(\bullet,\bullet)$ denotes the operator of mutual information. Based on the definition of mutual information, we have:
\begin{align*}
MI(\Delta P^{(S_i)}_j,Y^{(S_i)}_j)=H(Y^{(S_i)}_j)-H(Y^{(S_i)}_j|\Delta P^{(S_i)}_j),
\end{align*}
where $H(Y^{(S_i)}_j)$ represents the entropy of $Y^{(S_i)}_j$ and $H(Y^{(S_i)}_j|\Delta P^{(S_i)}_j)$ is the entropy of $Y^{(S_i)}_j$ conditioned on $\Delta P^{(S_i)}_j$. Since $H(Y^{(S_i)}_j)$ is constant, Equation \eqref{eq: maximize mutual information} is equivalent to minimizing the conditional entropy $H(Y^{(S_i)}_j|\Delta P^{(S_i)}_j)$, which can be expressed as:
\begin{align*}
\min\sum_{i=1}^M\sum_{j=1}^{\vert S_i\vert}  H(Y^{(S_i)}_j|\Delta P^{(S_i)}_j). 
\end{align*}
Due to the computational complexity of the conditional entropy $H(Y^{(S_i)}_j|\Delta P^{(S_i)}_j)$, it can be approximated by the cross-entropy between $f([\Delta P^{(S_i)}_j,X^{(S_i)}_j])$ and $Y^{(S_i)}_j$ \cite{ying2019gnnexplainer,luo2023time}, where $f([\Delta P^{(S_i)}_j,X^{(S_i)}_j])$ is the prediction obtained by concatenating $\Delta P^{(S_i)}_j$ and $X^{(S_i)}_j$ as an input to our proposed POND model, which will be illustrated in Section \ref{sec: discussion}. The fidelity loss is then expressed as:
\begin{align}
   \ell_{F}= \sum_{i=1}^M\sum_{j=1}^{\vert S_i\vert}Y^{(S_i)}_j \log f([\Delta P^{(S_i)}_j,X^{(S_i)}_j]).\label{eq:fidelity loss}
\end{align}
Now, we theoretically show that the learned prompt $\Delta P^{(S_i)}_j$, which minimizes the fidelity loss (\textit{i.e.}, Equation \eqref{eq:fidelity loss}), possesses the following properties:
\begin{property}[Preserving Fidelity]
    If $\Delta P^{(S_i)}_j$ minimizes Equation \eqref{eq:fidelity loss}, the mutual information between $\Delta P^{(S_i)}_j$ and the label $Y^{(S_i)}_j$ is equivalent to that between the time series input $X^{(S_i)}_j$ and the label $Y^{(S_i)}_j$, \textit{i.e.}, $MI(\Delta P^{(S_i)}_j,Y^{(S_i)}_j)=MI(X^{(S_i)}_j, Y^{(S_i)}_j)$.
    \label{pro:fidelity}
\end{property}
\begin{property}[Adding New Information] By minimizing Equation \eqref{eq:fidelity loss}, the generated prompt $\Delta P^{(S_i)}_j$ contains new information compared to the time series input $X^{(S_i)}_j$, \textit{i.e.}, $H(\Delta P^{(S_i)}_j)\geq H(X^{(S_i)}_j)$.
\label{pro:new information}
\end{property}
\indent Detailed proofs are provided in Section \ref{sec:property proof} in the Appendix. These properties demonstrate that minimizing Equation \eqref{eq:fidelity loss} ensures that the generated prompts will not decrease fidelity and may add new information to the time series input.\\
 \indent \textbf{High Distinction.} In addition to high fidelity, it is essential that the generated domain-specific prompt $\Delta P^{(S_i)}$ distinguishes the unique information of the source domain $S_i$ from other source domains. This unique information not only aids in understanding the differences between multiple time series source domains but also provides valuable insights for selecting suitable sources for domain adaptation. To achieve this, from the perspective of information theory, we define the objective to maintain high distinction as minimizing the mutual information of domain-specific prompts between different source domains, which should be minimized as follows:
\begin{align}
    \min \sum_{i_1\neq i_2} MI(\Delta P^{(S_{i_1})},\Delta P^{(S_{i_2})}), \label{eq:domain discrimination}
\end{align}
where $\Delta P^{( S_{i_1})}$ and $\Delta P^{( S_{i_2})}$ represent the domain-specific prompts of any two source domains $S_{i_1}$ and $S_{i_2}$. Equation \eqref{eq:domain discrimination} is computationally infeasible to minimize directly, but it can be achieved by minimizing the leave-one-out upper bound \cite{poole2019variational,luo2023time}. Other mutual information upper bounds, such as the contrastive log-ratio bound \cite{cheng2020club}, can also conveniently be incorporated into our framework. Therefore, the objective to encourage high distinction is formulated as minimizing the leave-one-out bound (\textit{i.e.}, discrimination loss):
\begin{align}
    \ell_D=\sum_{i_1\neq i_2}\mathbb{E}\log\frac{\exp( \text{sim}(\Delta P^{(S_{i_1})},\Delta P^{(S_{i_2})}))}{\sum_{i\neq i_1,i\neq i_2} \exp (\text{sim}(\Delta P^{(S_{i_1})},\Delta P^{(S_{i})}))},
    \label{eq:discrimination loss}
\end{align}
where $\text{sim}(\Delta P^{(S_{i_1})},\Delta P^{(S_{i_2})})=tr({(\Delta P^{(S_{i_1})})}^T\Delta P^{(S_{i_2})})$ denotes the inner product of the two domain-specific prompts $\Delta P^{(S_{i_1})}$ and $\Delta P^{(S_{i_2})}$, and $tr(A)$ represents the trace of any matrix $A$.
\subsection{The Learning Objective}
\label{sec:learning objective}

After introducing two criteria for selecting good prompts, we present our learning objective in this section.

Combining the fidelity loss $\ell_F$ in Equation \eqref{eq:fidelity loss} and the discrimination loss $\ell_D$ in Equation \eqref{eq:discrimination loss}, our learning objective is expressed as follows:
\begin{align}
    &\min\nolimits_{P,g^{(S_i)}} G(P,g^{(S_i)}) = \ell_R + \lambda_1 \ell_D + \lambda_2 \ell_F, \label{eq:learning objective}
\end{align}
where $\ell_R = \frac{1}{M}\sum_{i=1}^M \frac{1}{|S_i|}\sum_{j=1}^{|S_i|} R(f([P+\Delta P^{(S_i)}_j,X^{(S_i)}_j]),Y^{(S_i)}_j)$ is the training loss that measures the performance of prompt tuning. Here, $R(\cdot,\cdot)$ is the loss function, and $[P+\Delta P^{(S_i)}_j,X^{(S_i)}_j]$ is the concatenation of the overall prompt $P+\Delta P^{(S_i)}_j$ and the time series input $X^{(S_i)}_j$. Two tuning parameters $\lambda_1, \lambda_2>0$ control the trade-off among the training loss, the fidelity loss, and the discrimination loss.

To optimize Equation \eqref{eq:learning objective}, we need to enumerate all source domains, which may be inefficient and unscalable \cite{luo2023time}. To address this, we propose a simple yet effective learning algorithm based on the classic Reptile meta-learning framework \cite{nichol2018first}, which randomly picks a source domain each time and conducts standard steps of gradient descent without the need for calculating second derivatives. The learning process is outlined in Algorithm \ref{algo:prompt meta learning}. Specifically, Line 3 updates the prompt generator $g^{(S_\tau)}$, and Lines 4-5 update the common prompt $P$ through extrapolation. Here, the local learning rate $\eta$ performs the gradient descent step, and the global learning rate $\delta$ performs the extrapolation step.

\begin{algorithm}[H]
\small
\caption{Reptile-based meta-learning for Prompt Tuning}
\label{algo:prompt meta learning}
\begin{algorithmic}[1]
\REQUIRE $(X^{(S_i)}_j,Y^{(S_i)}_j)$, the global learning rate $\delta\in (0,1]$, the local learning rate $\eta>0$, the number of global steps $N$.
\ENSURE the common prompt $P$, the prompt generator $g^{(S_i)}$.\\
\FOR{$i=1$ to $N$}
\STATE Randomly pick a source time series domain $S_\tau$.
\STATE $g^{(S_\tau)} \leftarrow g^{(S_\tau)} - \eta \nabla_{g^{(S_\tau)}} G$.
\STATE $Q \leftarrow P - \eta \nabla_P \ell_T$.
\STATE $P \leftarrow P + \delta(Q - P)$.
\ENDFOR
\end{algorithmic}
\end{algorithm}

After learning the common prompt $P$ and the prompt generator $g^{(S_i)}$, they can be utilized for target domain transfer. Specifically, the prompt generator $g^{(T)}$ is optimized by the labeled time series pairs $(X^{(T)}_i,Y^{(T)}_i)$ in the target domain $T$ as follows:
\begin{align}
    \min\nolimits_{g^{(T)}} \frac{1}{|T|}\sum_{i=1}^{|T|} R(f([P+\Delta P^{(T)}_i,X^{(T)}_i]),Y^{(T)}_i), \label{eq:few shot transfer}
\end{align}
where $\Delta P^{(T)}_i = g^{(T)}(X^{(T)}_{i}) \in \mathbb{R}^{m \times n}$ is the instance-level domain-specific prompt of the time series input $X^{(T)}_{i}$, and the domain-level domain-specific prompt of the target domain $T$ is $\Delta P^{(T)} = \frac{1}{|T|} \sum_{j=1}^{|T|} \Delta P^{(T)}_j$. However, $g^{(T)}$ may not be reliable for prediction due to the limited labeled data involved. To handle this, $\Delta P^{(T)}$ is utilized as a heuristic to find the most similar source domain by the simple nearest neighbor rule (\textit{i.e.}, prompt adaptation):
\begin{align}
    S_i = \arg\max\nolimits_{S_i} \text{sim}(\Delta P^{(S_i)},\Delta P^{(T)}),
    \label{eq:nearest neighbor}
\end{align}
where $\text{sim}(\Delta P^{(S_i)},\Delta P^{(T)})$ is a similarity function (\textit{e.g.}, cosine similarity) between the domain-specific prompts $\Delta P^{(S_i)}$ and $\Delta P^{(T)}$. Then, we utilize the prompt generator $g^{(S_i)}$ for prediction in the target domain $T$: $f([P+g^{(S_i)}(X^{(T)}_j), X^{(T)}_j])$.
\subsection{Discussion}
\label{sec: discussion}
In this section, we discuss the model architecture and implementation, the theoretical aspects of our proposed POND model, and its comparison with previous papers.
\subsubsection{Model Architecture and Implementation}
For the model architecture of our proposed POND model, we employ the popular Mixture of Expert (MoE) technique to enhance performance \cite{fedus2022switch}: each expert makes an independent prediction, and the router is responsible for learning probability distributions over all predictions. The overall output of our POND model is a linear combination of all predictions.

For the architecture of a single expert, the time series input is fed into ``a patching layer'' (i.e., splitting a timeseries input into subseries-level patches \cite{nie2022time}), a projection layer, a position embedding layer, a transformer layer, and a linear head sequentially.

The model implementation is illustrated in the following steps: 
\begin{enumerate}
\item \textbf{Model Pretraining:} All experts of our POND model are pretrained by combining some labeled data from all source domains (e.g. $60\%$), and the router, which aggregates outputs from all experts to make final predictions, is pretrained using the same labeled data.
    \item \textbf{Prompt Tuning:} Given the pretrained POND model, other labeled time series data from all source domains (e.g. $40\%$) are utilized to learn the common prompt $P$ and the prompt generator $g^{(S_i)}$ by Equation \eqref{eq:learning objective} (\textit{i.e.}, Algorithm \ref{algo:prompt meta learning}), and the prompt generator of the target domain $g^{(T)}$ is optimized by Equation \eqref{eq:few shot transfer}.
    \item \textbf{Prompt Adaptation:} The most similar source domain is selected by Equation \eqref{eq:nearest neighbor}, whose prompt generator will be used in the target domain for prediction.
\end{enumerate}
\subsubsection{Theoretical Analysis}
 We demonstrate the commonality and differences of our proposed POND model compared with traditional prompt-tuning from the theoretical perspective. Specifically, we prove that our proposed POND model shares the universal approximation with prompt tuning, and then we illustrate that our proposed POND model overcomes the limitation of prompt tuning. Without loss of generality, we assume that only one expert model is available, and $\zeta$ is removed (\textit{i.e.}, $\Delta P^{(S_i)}_{j} = g^{(S_i)}(X^{(S_i)}_{j};\zeta)=g^{(S_i)}(X^{(S_i)}_{j})$). Proofs of all theorems below are shown in Section \ref{sec:theorem proof} in the Appendix due to space limitations.\\
\indent  One recent paper theoretically proves the universality of prompt tuning \cite{wang2023universality}, and it can be extended to our proposed POND model. Specifically, for any $\mathcal{L}$-Lipschitz function $\mathcal{F}: [0,1]^{n\times L}\rightarrow [0,1]^{\vert C\vert} $ under norm $q$, it satisfies the following: $\forall x_1,x_2\in [0,1]^{n\times L}, \Vert \mathcal{F}(x_1)-\mathcal{F}(x_2)\Vert_q\leq \mathcal{L}\Vert x_1-x_2\Vert_q$. The approximation error under $q$ norm is defined as $d_q(\mathcal{F}_1,\mathcal{F}_2)=(\int\Vert \mathcal{F}_1(x)-\mathcal{F}_2(x) \Vert^q_q dx)^{\frac{1}{q}}$. Then Theorem \ref{theo: universality} states that our proposed POND model can approximate any time series classifier, which are trained from specific source domains.
\begin{theorem}[Universality of our POND Model]
\label{theo:universality}
    Let $1\leq q<\infty$ and $\varepsilon>0$, and $\mathcal{F}^{(S_i)}: [0,1]^{n\times L}\rightarrow [0,1]^{\vert C\vert}$ is a time series classifer, which is trained from source domain $S_i$ and is $\mathcal{L}$-Lipschitz, there exist a prompt length $m$ and a POND model $f$ such that for any  $\mathcal{F}^{(S_i)}$,  we can find a domain-specific prompt generator $g^{(S_i)}: [0,1]^{n\times L}\rightarrow \mathbb{R}^{n\times m}$ from source domain $S_i$ with $d_q(f([P+g^{(S_i)}(\cdot),\cdot]),\mathcal{F}^{(S_i)})<\varepsilon$ for all $S_i(i=1,2,\cdots M)$.
    \label{theo: universality}
\end{theorem}
\indent Not only our proposed POND model shares the universality, it also overcomes the limitations of prompt tuning. The following theorem states that while prompt tuning may not be flexible enough to learn some labeled time series pairs, our proposed POND model can overcome this limitation.
\begin{theorem}[Flexibility of of our POND Model]
    Consider two labeled time series pairs $(X^{(S_1)}_1=[\mathcal{X}_1,\mathcal{X}_0],Y^{(S_1)}_1)$ and $(X^{(S_2)}_1=[\mathcal{X}_2,\mathcal{X}_0],Y^{(S_2)}_1)$  from two source domains $S_1$ and $S_2$, respectively, where $Y^{(S_1)}_1\neq Y^{(S_1)}_2$. For some proposed POND model $f$:\\
    (a).$[$The limitation of prompt tuning$]$There exists no prompt $P$ such that $f([P,X^{(S_i)}_1])=Y^{(S_i)}_1(i=1,2)$.\\
    (b).$[$Our POND model handles this limitation$]$ There exist the common prompt $P$ and the prompt generators $g^{(S_i)}(i=1,2)$ such that \\$f([P+g^{(S_i)}(X^{(S_i)}_1),X^{(S_i)}_1])=Y^{(S_i)}_1(i=1,2)$.
    \label{theo: flexibility}
\end{theorem}

\subsubsection{Comparison and Relation with Previous Methods}

Finally, we compare our proposed POND model with existing multi-source domain adaptation approaches, which have the following drawbacks:

(1). Neglection of domain-specific information. The common goal of existing methods is to make different domains indistinguishable \cite{zhao2020multi}. However, domain-specific information may be eliminated, which is important to select which source is the most similar to the target for adaptation. Our proposed POND model can address this by prompt tuning: the value of a prompt is updated by the gradient based on labeled data, which provides domain-specific information.

(2). Inability to capture time-dependent information. Most existing methods are designed to address domain adaptation problems in the fields of computer vision and NLP whose information is static, and they are not able to capture time-dependent information such as trends and distribution shifts of the time series. Our proposed POND model can address this by our proposed novel conditional module: it is learned to generate a prompt for each time series input, which is flexible to learn time-dependent information.

we show that several classic methods are special cases of our proposed POND model.\\
 \textbf{1. Generalization of Prompt Tuning.} Let $\lambda_1=\lambda_2=0$, and $g^{(S_i)}=0$, then our proposed POND model is reduced to the classic prompt tuning \cite{lester2021power}.\\
 \textbf{2. Generalization of Information Bottleneck.} Let $\lambda_1=0$ and $P=0$, then our proposed POND is reduced to the famous information bottleneck \cite{tishby2000information}.\\
 \textbf{3. Generalization of IDPG.} Let $\lambda_1=\lambda_2=0$, and $P=0$. Then our proposed POND model is reduced to Instance-Dependent Prompt Generation (IDPG) \cite{wu2022idpg}.
\section{Experiments}
 In this section, we employ four benchmark datasets to evaluate our proposed POND model in comparison with six state-of-the-art methods. All experiments were conducted on a Linux server equipped with an Intel(R) Xeon(R) Silver 4214 CPU and an NVIDIA GPU running version 510. More experiments are included in the supplementary materials \footnote{Link of supplementary materials: \url{https://github.com/xianggebenben/Junxiang_Wang.github.io/blob/master/supplementary_material/KDD2024/supp.pdf}. \label{supp}} due to space limitations.
 \begin{table}
\scriptsize
    \caption{Statistics of four datasets.}
    \centering
    \begin{tabular}{c|c|c|c|c|c|c}
    \hline\hline         
    Dataset& \tabincell{c}{\# Domain}& \tabincell{c}{\# Channel} & \tabincell{c}{\# Class}&  \tabincell{c}{Seq Len} & \tabincell{c}{\# Train} & \tabincell{c}{\# Test}  \\\hline
HAR&30&9&6&128&2300&990\\ \hline
WISDM&36&3&6&128&1350&720\\ \hline
HHAR&9&3&6&128&12716&5218\\ \hline
SSC&20&1&5&3000&14280&6130\\ \hline\hline
    \end{tabular}
    \label{tab:dataset}
    \vspace{-0.7cm}
\end{table}
\subsection{Experimental Settings}
\textbf{Benchmark Dataset:} We evaluated the performance of all methods on four benchmark datasets, HAR, WISDM, HHAR and SSC \cite{ragab2023adatime}. The statistics of all benchmark datasets are shown in Table \ref{tab:dataset}, which are introduced as follows:

1. HAR \cite{anguita2013public}: The Human Activity Recognition (HAR) dataset incorporates data collected from three sensors—accelerometer, gyroscope, and body sensors—deployed on 30 subjects (\textit{i.e.}, domains) engaged in six distinct activities.

2. WISDM \cite{kwapisz2011activity}: The WIreless Sensor Data Mining (WISDM) dataset, using accelerometer sensors, involves 36 subjects participating in activities similar to the HAR dataset, with additional challenges due to class distribution imbalances among different subjects.

3. HHAR \cite{stisen2015smart}: The Heterogeneity Human Activity Recognition (HHAR) dataset was collected from 9 subjects using sensor readings from smartphones and smartwatches.

4. SSC \cite{goldberger2000physiobank}: The Sleep Stage Classification (SSC) problem aims to categorize electroencephalography (EEG) signals into five stages. We utilize the Sleep-EDF dataset \cite{goldberger2000physiobank}, including EEG recordings from 20 healthy subjects.\\
\indent \textbf{Comparison Methods:} We compared our proposed POND method with six state-of-the-art time series domain adaptation approaches: Raincoat \cite{he2023domain}, CoDATs \cite{wilson2020multi}, Deep Coral \cite{sun2016deep}, MMDA \cite{rahman2020minimum}, DIRT-T \cite{shu2018dirt} and DSAN \cite{zhu2020deep}. All comparison methods are introduced as follows:

1. Raincoat \cite{he2023domain}: it is an unsupervised domain adaptation method addressing both feature and label shifts.

2. CoDATs \cite{wilson2020multi}: it is the first method to handle multi-source domain adaptation through adversarial training with weak supervision.

3. Deep Coral \cite{sun2016deep}: it minimizes domain shift by aligning second-order statistics of source and target distributions.

4. MMDA \cite{rahman2020minimum}: it integrates Maximum Mean Discrepancy (MMD) and CORrelation ALignment (CORAL) along with conditional entropy minimization to address domain shift.

5. DIRT-T \cite{shu2018dirt}: it utilizes adversarial training, conditional entropy, and a teacher model to align source and target domains.

6. DSAN \cite{zhu2020deep}: it minimizes the discrepancy between source and target domains via a Local Maximum Mean Discrepancy (LMMD) that aligns relevant subdomain distributions.

\indent \textbf{Metrics:} Two performance metrics were employed: Macro-F1 score and Accuracy. Macro-F1 is the unweighted mean of per-class F1 scores, treating all classes equally. Accuracy is the ratio of accurately predicted samples to all samples.\\
\indent\textbf{Hyperparameter Settings:} We adapted the setting of supervised domain adaptation, where ten samples in the target domain were used for domain transfer. All source-target scenarios were selected randomly to ensure the fairness of the performance evaluation. Single-source domain adaptation methods (e.g. Raincoat) were trained by combining all source domains. For the training set of all time series source domains, $60\%$ was used for pretraining our POND model, $20\%$ for prompt tuning, and $20\%$ for validation sets. The batch size was set to 16. The number of global steps $N$, global learning rate $\delta$ and the local learning rate $\eta$ were set to 50, 0.01 and 0.001, respectively. The number of experts was set to three.  The prompt generator is a two-layer Multi-Layer Perceptron (MLP) with Tanh activation. For the transformer model, the numbers of encoder layers, decoder layers, and heads in the multi-head attention were set to 2, 1, and 4, respectively. The dimensions of the multi-head attention and the feed-forward layer were set to 16 and 128, respectively. The hyperparameters $\lambda_1$ and $\lambda_2$ were chosen based on performance on the validation set. $\lambda_1$ and $\lambda_2$, along with other hyperparameters such as the number of epochs, are provided in Table \ref{tab:hyperparameter}. All methods were averaged by ten times.
\begin{table}[ht]
\vspace{-0.3cm}
\small
    \centering
        \caption{Hyperparameters of all datasets.}
        \vspace{-0.3cm}
    \begin{tabular}{c|c|c|c|c}
    \hline\hline
         Dataset&$\#$Epochs& Prompt Length & $\lambda_1$ & $\lambda_2$\\\hline
         HAR& 50&5&1&1\\ \hline
         WISDM&200&3&1&1 \\ \hline
         HHAR&200&5&1&1\\ \hline
         SSC&100&10&0.1&0.1\\ \hline\hline
    \end{tabular}
    \vspace{-0.3cm}
    \label{tab:hyperparameter}
\end{table}
\subsection{Experimental Results}
\begin{figure}
\begin{minipage}{0.49\linewidth}
   \centering
    \includegraphics[width=\linewidth]{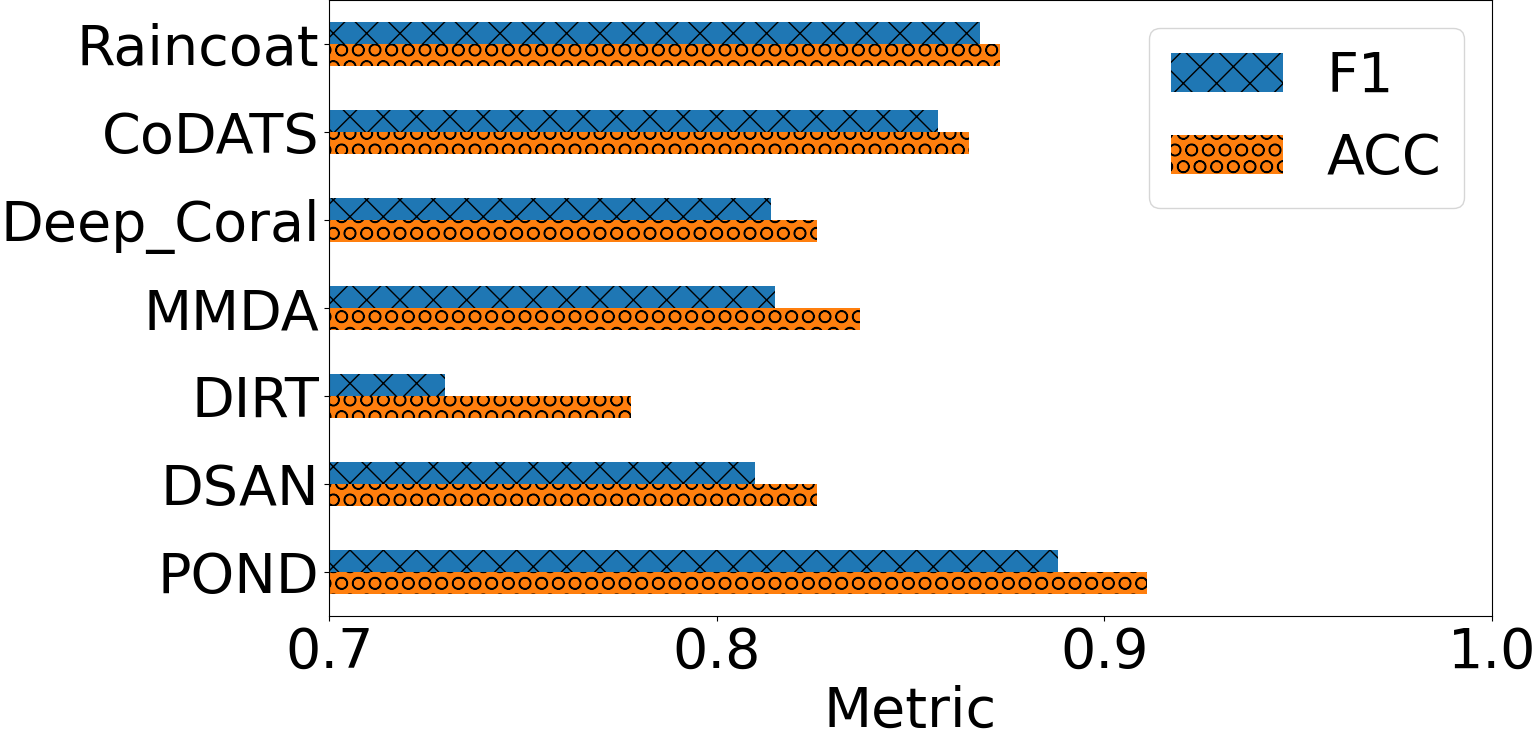}
    \centerline{(a). HAR.}
\end{minipage}
    \hfill
\begin{minipage}{0.49\linewidth}
   \centering
    \includegraphics[width=\linewidth]{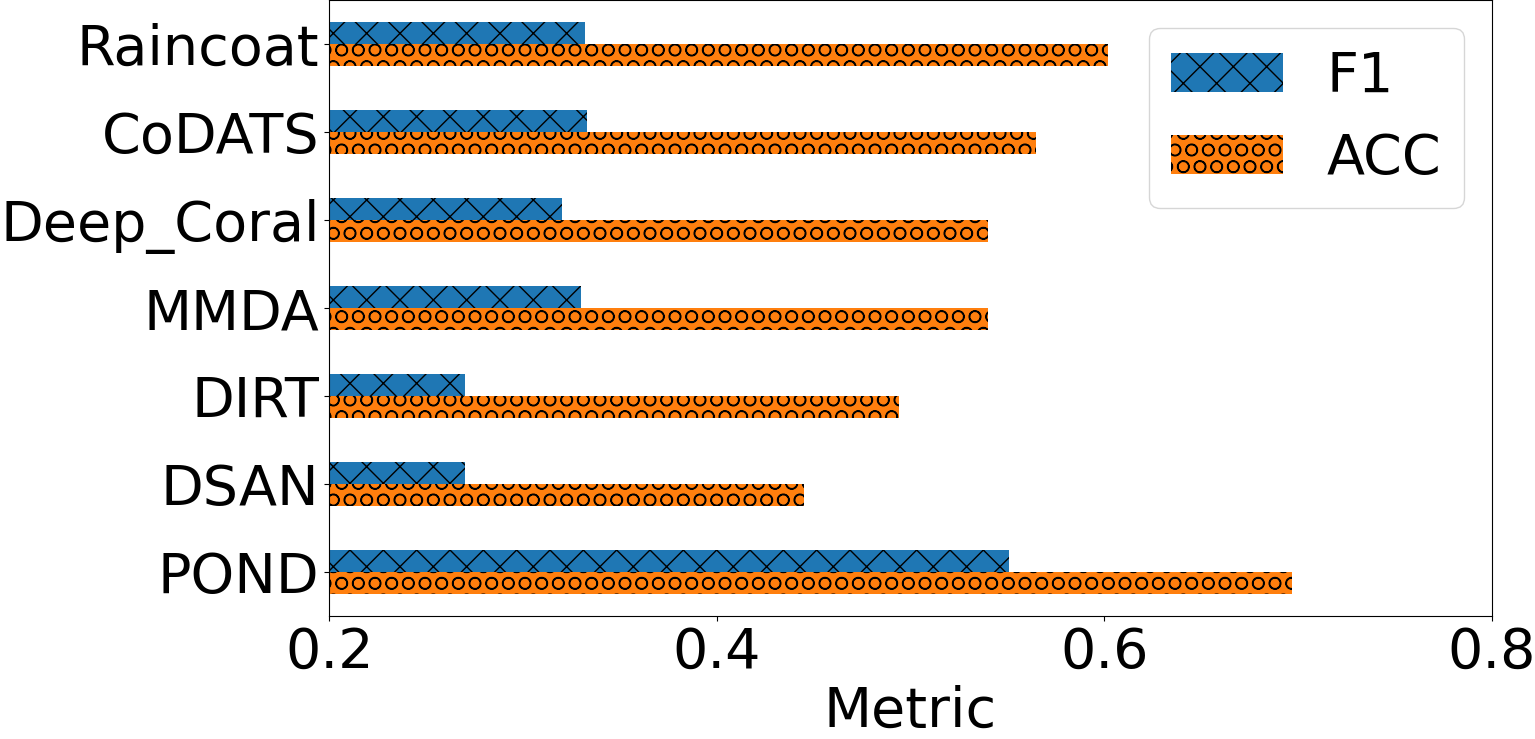}
    \centerline{(b). WISDM.}
\end{minipage}    
\vfill
\begin{minipage}{0.49\linewidth}
  \centering
    \includegraphics[width=\linewidth]{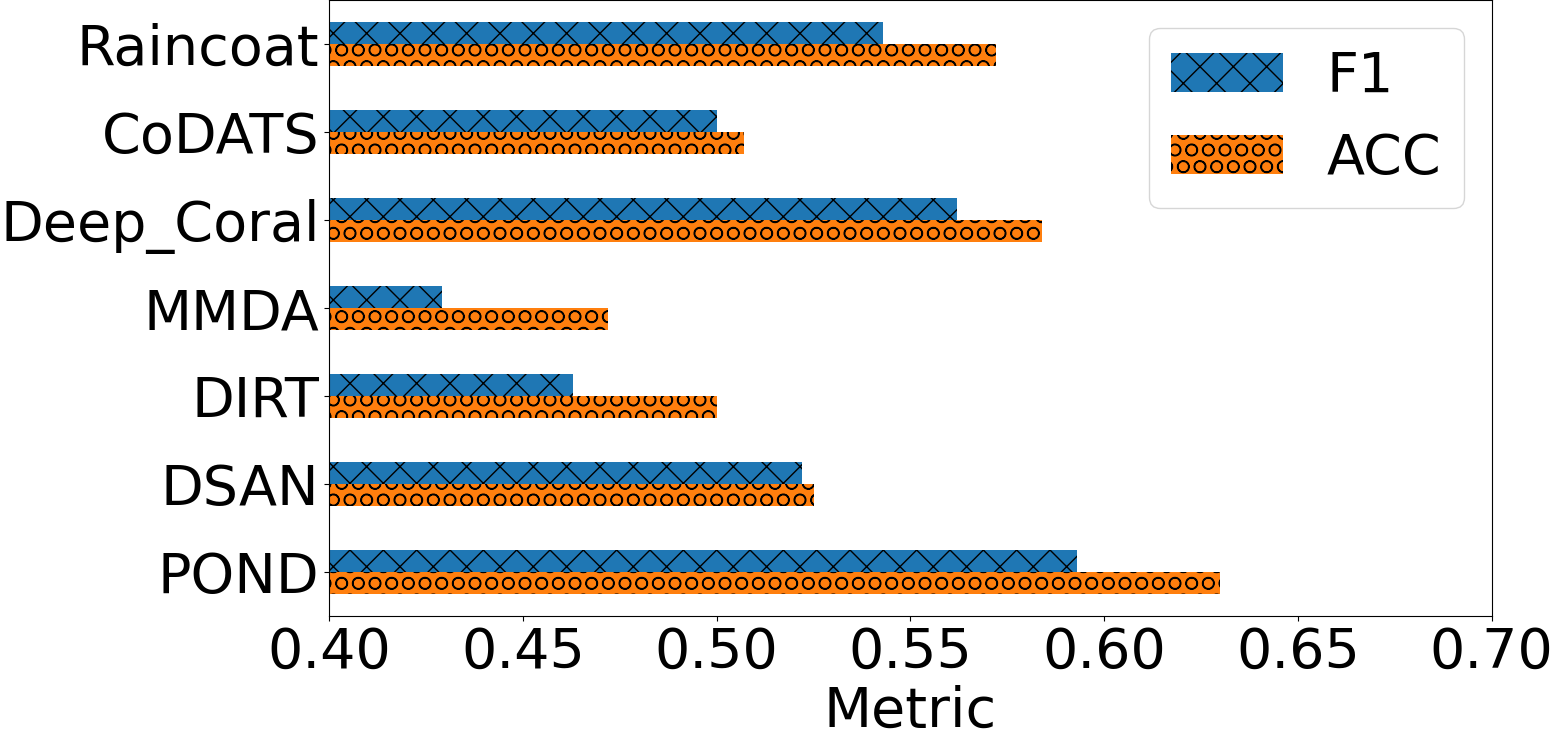}
    \centerline{(c). HHAR.}
\end{minipage}
\hfill
\begin{minipage}{0.49\linewidth}
  \centering
    \includegraphics[width=\linewidth]{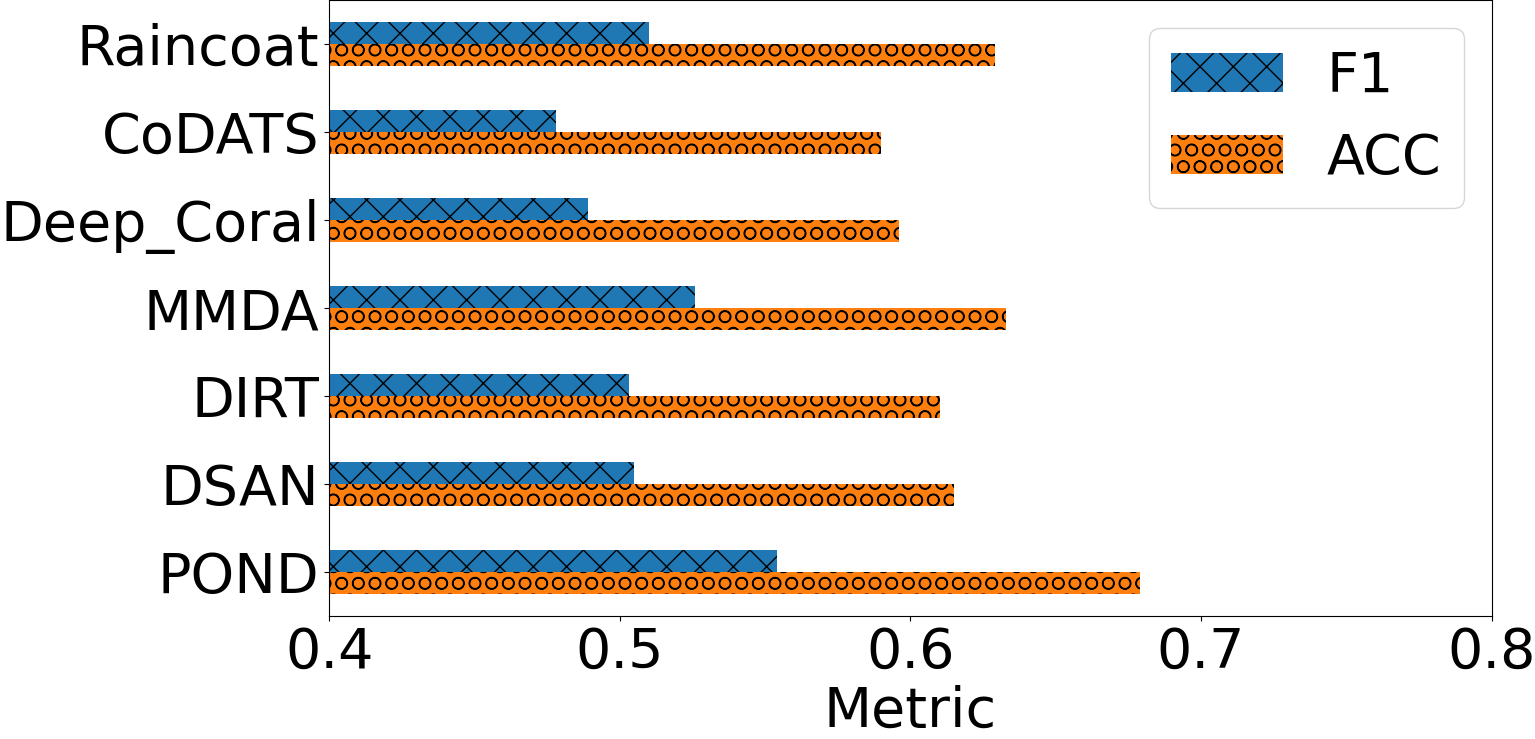}  
         \centerline{(d). SSC.}
\end{minipage}
\vspace{-0.2cm}
\caption{The F1-score and accuracy of all methods on four benchmark datasets: the proposed POND outperforms comparison methods consistently.}
\label{fig:performance bar}
\vspace{-0.5cm}
\end{figure}
\textbf{Performance Evaluation:} We conducted a comprehensive performance evaluation to test all methods across approximately 50 scenarios on four datasets. Figure \ref{fig:performance bar} displays the F1-score and accuracy of all methods on these datasets. Our proposed POND method consistently outperforms others across all four datasets. Specifically, on the HAR dataset, the F1-score of POND is approximately 0.9, only $2\%$ lower than the top-performing comparison method, Raincoat. The F1-score gaps on the HHAR, and SSC datasets are $5\%$ and $4.4\%$, respectively. The largest gap is observed in the WISDM dataset, where the F1-score and accuracy of POND hover around $0.6$ and $0.7$, while all comparison methods score below 0.35 and 0.6, respectively. Considering the inherent difficulty of training on the WISDM dataset due to class imbalance, this highlights the effectiveness of our proposed POND, especially on challenging datasets. \\
 \indent Among the comparison methods, Raincoat emerges as the best overall. In terms of F1-score, Raincoat outperforms MMDA by $5\%$ on the HAR dataset and shows an $8\%$ superiority over CoDATs on the HHAR dataset. For accuracy, Raincoat performs $7\%$ better than DIRT on the HHAR dataset and surpasses Deep Coral by $3\%$ on the SSC dataset. CoDATs and Deep Coral also demonstrate competitive performance, achieving around $55\%$ accuracy on the WISDM dataset, while DSAN lags behind at $45\%$. On the other hand, MMDA, DIRT, and DSAN exhibit varying performance across datasets. For instance, DSAN performs comparably to Raincoat on the SSC dataset but ranks the lowest on the WISDM dataset.\\
\begin{table*}[ht]
    \caption{F1-score on different scenarios of four datasets: the proposed POND model outperforms all comparison methods.}
    \vspace{-0.3cm}
\small
    \centering
    \begin{tabular}{l|c|c|c|c|c|c|c|c}
    \hline\hline
        Scenario & Raincoat &  CoDATs &  Deep\_Coral  & MMDA & DIRT & DSAN &  POND & Target Only \\ \hline
        HAR 1-15 $\rightarrow$ 16 & 0.823 $\pm$ 0.094 & 0.767 $\pm$ 0.093 & 0.773 $\pm$ 0.082 & 0.679 $\pm$ 0.084 & 0.612 $\pm$ 0.135 & 0.738 $\pm$ 0.095 & \textbf{0.849 $\pm$ 0.021} & 0.856 $\pm$ 0.027 \\ \hline
        HAR 1-15 $\rightarrow$ 20 & 0.872 $\pm$ 0.142 & 0.932 $\pm$ 0.025 & 0.923 $\pm$ 0.023 & 0.921 $\pm$ 0.034 & 0.848 $\pm$ 0.101 & 0.929 $\pm$ 0.033 & \textbf{0.968 $\pm$ 0.021} & 0.983 $\pm$ 0.018 \\ \hline
        HAR 1-15 $\rightarrow$ 21 & 0.867 $\pm$ 0.141 & 0.903 $\pm$ 0.070 & 0.882 $\pm$ 0.028 & \textbf{0.974 $\pm$ 0.039} & 0.921 $\pm$ 0.090 & 0.909 $\pm$ 0.110 & 0.972 $\pm$ 0.021 & 1.000 $\pm$ 0.000 \\ \hline
        HAR 1-15 $\rightarrow$ 28 & 0.766 $\pm$ 0.107 & 0.775 $\pm$ 0.166 & \textbf{0.852 $\pm$ 0.044} & 0.778 $\pm$ 0.085 & 0.671 $\pm$ 0.175 & 0.783 $\pm$ 0.046 & 0.829 $\pm$ 0.018 & 0.853 $\pm$ 0.019 \\ \hline
         HAR 16-20 $\rightarrow$ 1 & 0.792 $\pm$ 0.072 & 0.744 $\pm$ 0.053 & 0.667 $\pm$ 0.077 & 0.654 $\pm$ 0.074 & 0.546 $\pm$ 0.060 & 0.698 $\pm$ 0.037 & \textbf{0.883 $\pm$ 0.017} & 0.986 $\pm$ 0.010 \\ \hline
        HAR 16-20 $\rightarrow$ 2 & 0.825 $\pm$ 0.048 & 0.821 $\pm$ 0.151 & 0.796 $\pm$ 0.055 & 0.651 $\pm$ 0.045 & 0.509 $\pm$ 0.050 & 0.652 $\pm$ 0.057 & \textbf{0.936 $\pm$ 0.017} & 0.943 $\pm$ 0.024 \\ \hline
        HAR 16-20 $\rightarrow$ 3 & 0.814 $\pm$ 0.028 & 0.746 $\pm$ 0.078 & 0.741 $\pm$ 0.058 & 0.657 $\pm$ 0.033 & 0.605 $\pm$ 0.056 & 0.565 $\pm$ 0.043 & \textbf{0.878 $\pm$ 0.018} & 0.978 $\pm$0.013 \\ \hline
        HAR 16-20 $\rightarrow$ 4 & 0.679 $\pm$ 0.084 & 0.605 $\pm$ 0.082 & 0.479 $\pm$ 0.110 & 0.513 $\pm$ 0.058 & 0.336 $\pm$ 0.110 & 0.436 $\pm$ 0.032 & \textbf{0.754 $\pm$ 0.033} & 0.921 $\pm$ 0.018 \\ \hline
        WISDM 0-17 $\rightarrow$ 18 & 0.379 $\pm$ 0.061 & 0.384 $\pm$ 0.049&  0.346 $\pm$ 0.023 & 0.297 $\pm$ 0.016 & 0.300 $\pm$ 0.041 & 0.287 $\pm$ 0.045 & \textbf{0.606 $\pm$ 0.020} & 0.705 $\pm$ 0.046 \\ \hline
        WISDM 0-17 $\rightarrow$ 20 & 0.354 $\pm$ 0.040 & 0.368 $\pm$ 0.039 & 0.376 $\pm$ 0.031 & 0.452 $\pm$ 0.098 & 0.347 $\pm$ 0.071 & 0.269 $\pm$ 0.064 & \textbf{0.570 $\pm$ 0.023} & 0.704 $\pm$ 0.051 \\ \hline
        WISDM 0-17 $\rightarrow$ 21 & 0.355 $\pm$ 0.057 & 0.310 $\pm$ 0.088 & 0.259 $\pm$ 0.018 & 0.250 $\pm$ 0.000 & 0.276 $\pm$ 0.055 & 0.245 $\pm$ 0.046 & \textbf{0.450 $\pm$ 0.026} & 0.636 $\pm$ 0.095 \\ \hline
        WISDM 0-17 $\rightarrow$ 23 & 0.306 $\pm$ 0.015 & 0.327 $\pm$ 0.075 & 0.318 $\pm$ 0.031 & 0.327 $\pm$ 0.023 & 0.271 $\pm$ 0.016 & 0.277 $\pm$ 0.044 & \textbf{0.482 $\pm$ 0.017} & 0.538 $\pm$ 0.034 \\ \hline
        WISDM 0-17 $\rightarrow$ 25 & 0.365 $\pm$ 0.030 & 0.540 $\pm$ 0.125 & 0.435 $\pm$ 0.043 & 0.436 $\pm$ 0.094 & 0.314 $\pm$ 0.107 & 0.353 $\pm$ 0.120 & \textbf{0.559 $\pm$ 0.050} & 0.672 $\pm$ 0.039 \\ \hline
        WISDM 0-17 $\rightarrow$ 28 & 0.399 $\pm$ 0.028 & 0.431 $\pm$ 0.033 & 0.418 $\pm$ 0.032 & 0.454 $\pm$ 0.064 & 0.304 $\pm$ 0.044 & 0.339 $\pm$ 0.030 & \textbf{0.656 $\pm$ 0.046} & 0.689 $\pm$ 0.048 \\ \hline
        WISDM 0-17 $\rightarrow$ 30 & 0.314 $\pm$ 0.020 & 0.305 $\pm$ 0.028 & 0.298 $\pm$ 0.023 & 0.359 $\pm$ 0.072 & 0.266 $\pm$ 0.035 & 0.246 $\pm$ 0.076 & \textbf{0.670 $\pm$ 0.039} & 0.791 $\pm$ 0.028 \\ \hline
        WISDM 18-23 $\rightarrow$ 5 & 0.648 $\pm$ 0.001 & 0.558 $\pm$ 0.129 & 0.534 $\pm$ 0.102 & 0.510 $\pm$ 0.020 & 0.549 $\pm$ 0.097 & 0.484 $\pm$ 0.055 & \textbf{0.652 $\pm$ 0.035} & 0.734 $\pm$ 0.095 \\ \hline
        WISDM 18-23 $\rightarrow$ 6 & 0.544 $\pm$ 0.074 & 0.565 $\pm$ 0.143 & 0.437 $\pm$ 0.078 & 0.543 $\pm$ 0.160& 0.405 $\pm$ 0.089 & 0.454 $\pm$ 0.112 & \textbf{0.628 $\pm$ 0.033} & 0.872 $\pm$ 0.049 \\ \hline
        WISDM 18-23 $\rightarrow$ 7 & 0.588 $\pm$ 0.070 & 0.404 $\pm$ 0.117 & 0.530 $\pm$ 0.094 & 0.477 $\pm$ 0.060 & 0.518 $\pm$ 0.120 & 0.476 $\pm$ 0.127 & \textbf{0.672 $\pm$ 0.029} & 0.888 $\pm$ 0.035 \\ \hline
        HHAR 0-6 $\rightarrow$ 7 & 0.765 $\pm$ 0.142 & 0.652 $\pm$ 0.108 & 0.815 $\pm$ 0.105 & 0.641 $\pm$ 0.050 & 0.649 $\pm$ 0.005 & 0.730 $\pm$ 0.164 & \textbf{0.834 $\pm$ 0.014} & 0.861 $\pm$ 0.016 \\ \hline
        HHAR 5-8 $\rightarrow$ 2 & 0.321 $\pm$ 0.023 & 0.347 $\pm$ 0.082 & 0.309 $\pm$ 0.032 & 0.216 $\pm$ 0.032 & 0.276 $\pm$ 0.021 & 0.314 $\pm$ 0.095 & \textbf{0.352 $\pm$ 0.014} & 0.881 $\pm$ 0.018 \\ \hline
        SSC 0-9 $\rightarrow$ 16 & 0.578 $\pm$ 0.028 & 0.510 $\pm$ 0.044 & 0.537 $\pm$ 0.024 & 0.559 $\pm$ 0.027 & 0.523 $\pm$ 0.019 & 0.515 $\pm$ 0.044 & \textbf{0.568 $\pm$ 0.012} & 0.601 $\pm$ 0.018 \\ \hline
        SSC 0-9 $\rightarrow$ 17 & 0.511 $\pm$ 0.024 & 0.413 $\pm$ 0.118 & 0.452 $\pm$ 0.077 & 0.504 $\pm$ 0.060 & 0.530 $\pm$ 0.053 & 0.463 $\pm$ 0.081 & \textbf{0.559 $\pm$ 0.006} & 0.602 $\pm$ 0.014 \\ \hline
        SSC 0-9 $\rightarrow$ 18 & \textbf{0.605 $\pm$ 0.016} & 0.548 $\pm$ 0.037 & 0.544 $\pm$ 0.046 & 0.597 $\pm$ 0.032 & 0.574 $\pm$ 0.021 & 0.569 $\pm$ 0.046 & 0.604 $\pm$ 0.014 & 0.602 $\pm$ 0.013 \\ \hline
        SSC 0-9 $\rightarrow$ 19 & 0.562  $\pm$ 0.024 & 0.540 $\pm$ 0.052 & 0.531 $\pm$ 0.055 & \textbf{0.570 $\pm$ 0.044} & 0.565 $\pm$ 0.028 & 0.568 $\pm$ 0.080 & \textbf{0.570 $\pm$ 0.010} & 0.613 $\pm$ 0.019 \\ \hline
        SSC 10-12 $\rightarrow$ 8 & 0.294 $\pm$ 0.028 & 0.380 $\pm$ 0.066 & 0.379 $\pm$ 0.076 & 0.398 $\pm$ 0.060 & 0.322 $\pm$ 0.048 & 0.411 $\pm$ 0.046 & \textbf{0.470 $\pm$ 0.010} & 0.531 $\pm$ 0.019 \\ \hline\hline
        
    \end{tabular}
    \label{tab:performance}
    \vspace{-0.2cm}
\end{table*}
\indent Table \ref{tab:performance} presents the performance of all methods across various scenarios in four datasets, including the upper bound achieved by training and testing on the target domain. The reported values include means and standard deviations from ten implementations, with the best results highlighted in bold. The complete performance evaluation is available in the supplementary materials \footref{supp}. Overall, our proposed POND model consistently outperforms all methods, aligning with the observations in Figure \ref{fig:performance bar}. Notably, POND exhibits superior performance on the challenging WISDM dataset, as indicated by Figure \ref{fig:performance bar}. For instance, POND outperforms all comparison methods by at least $23\%$ when transferring from domains 0-17 to domain 18. While POND excels overall, there are instances where comparison methods outperform it. For example, Deep Coral performs better than POND by $2\%$ when transferring domains 1-15 to domain 28 on the HAR dataset, and MMDA marginally outperforms POND when transferring domains 1-15 to domain 21 on the HAR dataset. \\
 \indent In addition to superior performance, our proposed POND model demonstrates greater stability compared to all comparison methods, as indicated by lower standard deviations. For instance, the standard deviation of POND is $0.006$ when transferring domains 0-9 to domain 17 on the SSC dataset, while the standard deviations of all comparison methods range between $0.024$ and $0.118$, being at least 3 times larger than that of POND. Importantly, POND achieves results close to the upper bound in many scenarios, such as "HAR 1-15 $\rightarrow$ 16", "SSC 0-9 $\rightarrow $ 18", and "HHAR 0-6 $\rightarrow $ 7".
 \begin{table*}[]
 \caption{Ablation study on the WISDM dataset: all components of our proposed POND model contribute to the outstanding performance.}
 \vspace{-0.3cm}
 \scriptsize
\begin{tabular}{c|c|c|c|c|c|c|c|c|c}
\hline\hline
MoE & \tabincell{c}{Common\\Prompt} & \tabincell{c}{Prompt\\Generator} & 0-17$\rightarrow$ 22 & 0-17$\rightarrow$ 23 & 0-17$\rightarrow$ 24 & 0-17$\rightarrow$ 25 & 18-23$\rightarrow$ 5 & 18-23$\rightarrow$ 6& Overall \\\hline
\usym{2717}  & $\checkmark$           & \usym{2717}               & 0.622$\pm$0.057                                     & 0.415$\pm$0.015                                     & 0.510$\pm$0.030                                     & 0.581$\pm$0.036                                     & 0.623$\pm$0.058                                     & 0.516$\pm$0.038&0.545$\pm$0.039                                     \\\hline
\usym{2717}  & \usym{2717}            & $\checkmark$              & 0.646$\pm$0.064                                     & 0.396$\pm$0.048                                     & 0.527$\pm$0.030                                     & 0.573$\pm$0.034                                     & 0.628$\pm$0.051                                     & 0.512$\pm$0.057&0.547$\pm$0.047                                     \\\hline
\usym{2717}  & $\checkmark$           & $\checkmark$              & 0.632$\pm$0.069                                     & 0.384$\pm$0.041                                     & 0.498$\pm$0.032                                     & 0.572$\pm$0.045                                     & 0.611$\pm$0.055                                     & 0.514$\pm$0.025&  0.535$\pm$0.045                                 \\\hline
$\checkmark$ & \usym{2717}            & $\checkmark$              & 0.575$\pm$0.043                                     & 0.349$\pm$0.029                                     & 0.517$\pm$0.032                                     & 0.584$\pm$0.030                                    & 0.621$\pm$0.056                                     & 0.578$\pm$0.035&0.537$\pm$0.038                                     \\\hline
$\checkmark$ & $\checkmark$           & \usym{2717}               & 0.719$\pm$0.062                                     & 0.405$\pm$0.052                                     & 0.529$\pm$0.042                                     & 0.588$\pm$0.034                                     & 0.616$\pm$0.050                                     & 0.565$\pm$0.049&0.570$\pm$0.048                                     \\\hline
$\checkmark$ & $\checkmark$           & $\checkmark$              & \textbf{0.725$\pm$0.031}                                     & \textbf{0.482$\pm$0.017}                                     & \textbf{0.559$\pm$0.050}                                     & \textbf{0.695$\pm$0.035}                                     & \textbf{0.652$\pm$0.035}                                     & \textbf{0.628$\pm$0.033}&\textbf{0.624$\pm$0.034}\\\hline\hline                                  
\end{tabular}
\label{tab:ablation study}
\vspace{-0.3cm}
\end{table*}
\\
\indent \textbf{Ablation Study}: Next, we demonstrate the ablation study of the proposed POND method, whose goal is to identify whether all components of our proposed POND model contribute to the performance. Specifically, we explore the necessity of the  MoE technique, common prompt, and prompt generator. The challenging WISDM dataset was utilized to test the performance. Table \ref{tab:ablation study} illustrates the performance of different scenarios, all of which were averaged by $10$ times. The first two rows show the performance with the common prompt, and the prompt generator available only, respectively. The fourth to sixth rows demonstrate the performance without the MoE, common prompt, and prompt generator, respectively, and the last row shows the performance of the complete POND model. Overall, our proposed POND model performs best when the MoE, common prompt, and prompt generator are all available, which suggests that all components are necessary for the outstanding performance of our proposed POND model. For example,  in the scenario of “18-23$\rightarrow$ 6”, the best performance without any component only achieves a performance no more than 0.58, whereas that of the complete POND model is $5\%$ better. The gap is widened to $7\%$ for the scenario “0-17$\rightarrow$ 25”.\\ 
\begin{figure}
\begin{minipage}{0.45\linewidth}
   \centering
    \includegraphics[width=\linewidth]{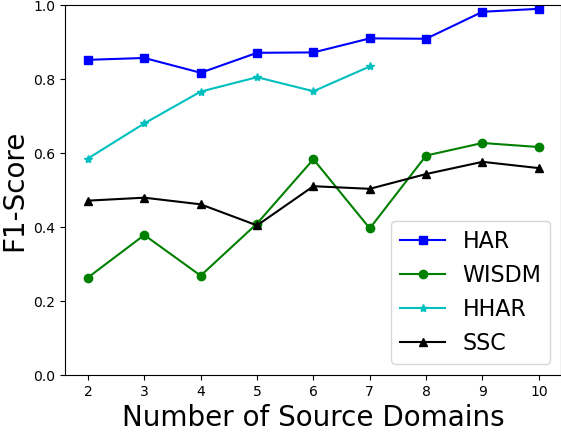}
    \centerline{(a). F1-score.}
\end{minipage}
    \hfill
\begin{minipage}{0.45\linewidth}
   \centering
    \includegraphics[width=\linewidth]{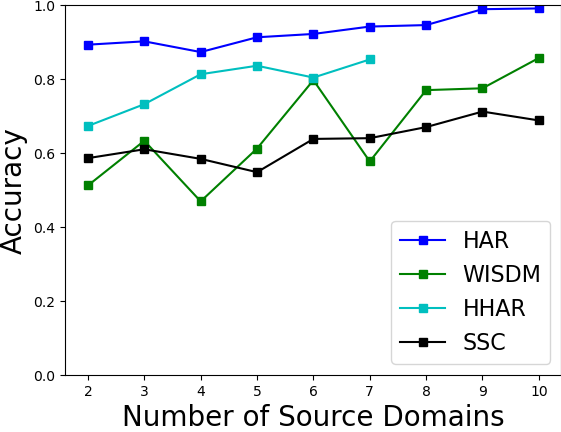}
    \centerline{(b). Accuracy.}
\end{minipage}
\vspace{-0.2cm}
\caption{The F1-score and accuracy of the proposed POND model with different source domains: the performance grows with the increase of source domains. (The HHAR dataset has less than 10 domains.)}
\vspace{-0.5cm}
\label{fig:different source domains}
\end{figure}
\indent \textbf{Sensitivity Analysis:} In this section, we explore how source domains influence performance on the target domain. Figure \ref{fig:different source domains} illustrates the relationship between performance metrics (F1-score and accuracy) and the number of source domains, averaged over 10 implementations. Generally, our proposed POND model demonstrates improved performance with an increasing number of source domains. For instance, POND achieves $50\%$ accuracy with two source domains for training, but this figure rises by $30\%$ when an additional 8 source domains are included. Similarly, the F1-score of POND increases by $20\%$ when the number of source domains changes from $2$ to $6$. However, some exceptions exist. For example, there is a notable $25\%$ drop in F1-score when increasing the number of source domains from 6 to 7 on the WISDM dataset. Another instance involves a $5\%$ performance drop when increasing the source domains from $4$ to $5$ on the SSC dataset.\\
\begin{figure}
\begin{minipage}{0.45\linewidth}
   \centering
    \includegraphics[width=\linewidth]{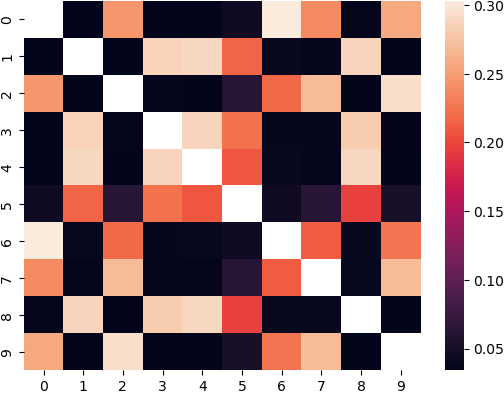}
    \centerline{(a). HAR.}
\end{minipage}
    \hfill
\begin{minipage}{0.45\linewidth}
   \centering
    \includegraphics[width=\linewidth]{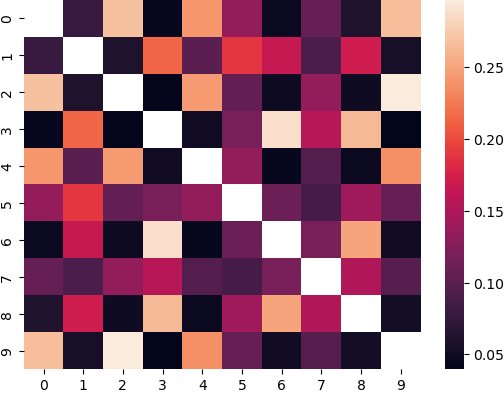}
    \centerline{(b). WISDM.}
\end{minipage}    
\vfill
\begin{minipage}{0.45\linewidth}
  \centering
    \includegraphics[width=\linewidth]{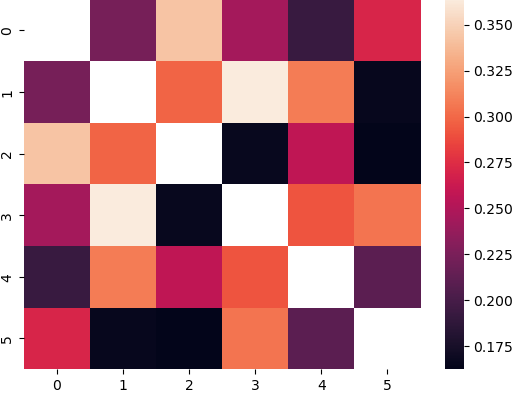}
    \centerline{(c). HHAR.}
\end{minipage}
\hfill
\begin{minipage}{0.45\linewidth}
  \centering
    \includegraphics[width=\linewidth]{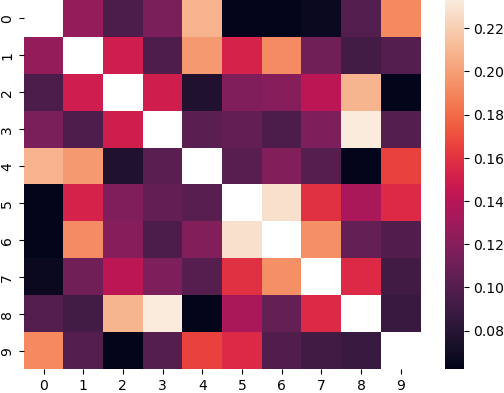}  
         \centerline{(d). SSC.}
\end{minipage}
\vspace{-0.3cm}
\caption{The visualization of  the exponent of discrimination loss: most pairs of source domains are well discriminated. }
\label{fig:visualization}
\vspace{-0.3cm}
\end{figure}
\indent \textbf{Visualization of Discrimination Loss:} Finally, we present a visualization of the discrimination loss $\ell_D$ for pairwise source domains. Figure \ref{fig:visualization} illustrates the exponents of discrimination losses for all pairs of source domains across four datasets. Both the X-axis and Y-axis represent the indexes of source domains. Darker colors indicate smaller discrimination losses, reflecting better domain discrimination. The diagonals are left blank. Overall, our proposed POND model effectively discriminates most source domains, as evidenced by the predominance of dark squares. For instance, domains 3-5 and domains 6-7 exhibit clear discrimination with losses below 0.05. Similar effective discrimination is observed for domain pairs 6 and 0 on the WISDM dataset, domain pairs 1 and 5 on the HHAR dataset, and domains 5-7 and 0 in the SSC dataset. However, discrimination losses for some domain pairs are larger than others. For instance, on the HAR dataset, the discrimination loss between domains 0 and 6 is the largest, approximately 0.30, but still within an acceptable range. It's worth noting that domain discrimination may not adhere to the transitive property. For example, domains 3 and 9, as well as domains 4 and 9, are well-discriminated, but domains 3 and 4 are relatively poor-discriminated.
\section{Conclusion}
 Time series domain adaptation is an important problem with wide-ranging applications. Existing techniques primarily address single-source domain adaptation, yet exploring adaptation from multiple domains holds promise for greater improvements. In this paper, we introduce POND, the first framework to utilize prompts for time series domain adaptation. We extend prompt tuning to time series analysis to capture common and domain-specific information from all source domains, introduce conditional modules for prompt generation, and propose criteria for selecting effective prompts. Through extensive experiments across 50 scenarios on four datasets, we demonstrate the efficacy and robustness of POND, outperforming all state-of-the-art methods by up to $66\%$ on the F1-score.
\bibliographystyle{acm}
\balance
\bibliography{example_paper}
\setcounter{property}{0}
\setcounter{theorem}{0}
\appendix
\textbf{Appendix}\\
\section{Mathematical Proofs}
\subsection{Proofs of Properties \ref{pro:fidelity} and \ref{pro:new information}}
\label{sec:property proof}
\begin{property}[Preserving Fidelity]
    If $\Delta P^{(S_i)}_j$ minimizes Equation \eqref{eq:fidelity loss}, the mutual information between $\Delta P^{(S_i)}_j$ and the label $Y^{(S_i)}_j$ is equivalent to that between the time series input $X^{(S_i)}_j$ and the label $Y^{(S_i)}_j$, \textit{i.e.}, $MI(\Delta P^{(S_i)}_j,Y^{(S_i)}_j)=MI(X^{(S_i)}_j, Y^{(S_i)}_j)$.
\end{property}
\begin{proof}
    From the definition of mutual information, we have:
    \begin{align*}
    &MI(\Delta P^{(S_i)}_j,Y^{(S_i)}_j)\\&=H(Y^{(S_i)}_j)-H(Y^{(S_i)}_j|\Delta P^{(S_i)}_j)\\&=H(Y^{(S_i)}_j)+\sum_{\Delta P^{(S_i)}_j,Y^{(S_i)}_j}p(\Delta P^{(S_i)}_j,Y^{(S_i)}_j)\log \frac{p(\Delta P^{(S_i)}_j,Y^{(S_i)}_j)}{p(\Delta P^{(S_i)}_j)}\\&=H(Y^{(S_i)}_j)\!+\!\sum_{X^{(S_i)}_j,Y^{(S_i)}_j}\sum_{\Delta P^{(S_i)}_j\in \mathbb{V}(X^{(S_i)}_j)}p(\Delta P^{(S_i)}_j,Y^{(S_i)}_j)\\&\log \frac{p(\Delta P^{(S_i)}_j,Y^{(S_i)}_j)}{p(\Delta P^{(S_i)}_j)}.
    \end{align*}
    where $\mathbb{V}(X^{(S_i)}_j)$ is the set of generated prompts of a time series input $X^{(S_i)}_j$. Because $\Delta P^{(S_i)}_j$ is a function of $X^{(S_i)}_j$ only, this means that $p(\Delta P^{(S_i)}_j|X^{(S_i)}_j,Y^{(S_i)}_j)=p(\Delta P^{(S_i)}_j|X^{(S_i)}_j)$. Since the mapping between $X^{(S_i)}_j$ and $\Delta P^{(S_i)}_j$ is one to many, for each $\Delta P^{(S_i)}_j\in \mathbb{V}(X^{(S_i)}_j)$, we have $p(\Delta P^{(S_i)}_j,Y^{(S_i)}_j)=p(\Delta P^{(S_i)}_j,X^{(S_i)}_j,Y^{(S_i)}_j)$, and $p(\Delta P^{(S_i)}_j)=p(\Delta P^{(S_i)}_j|X^{(S_i)}_j)p(X^{(S_i)}_j)$. Therefore, we have
    \begin{align*}
        \frac{p(\Delta P^{(S_i)}_j,Y^{(S_i)}_j)}{p(\Delta P^{(S_i)}_j)}&=\frac{p(\Delta P^{(S_i)}_j,X^{(S_i)}_j,Y^{(S_i)}_j)}{p(\Delta P^{(S_i)}_j|X^{(S_i)}_j)p(X^{(S_i)}_j)}\\&=\frac{p(\Delta P^{(S_i)}_j|X^{(S_i)}_j,Y^{(S_i)}_j)p(X^{(S_i)}_j,Y^{(S_i)}_j)}{p(\Delta P^{(S_i)}_j|X^{(S_i)}_j)p(X^{(S_i)}_j)}\\&=\frac{p(\Delta P^{(S_i)}_j|X^{(S_i)}_j)p(X^{(S_i)}_j,Y^{(S_i)}_j)}{p(\Delta P^{(S_i)}_j|X^{(S_i)}_j)p(X^{(S_i)}_j)}\\&=\frac{p(X^{(S_i)}_j,Y^{(S_i)}_j)}{p(X^{(S_i)}_j)}.
    \end{align*}
    \begin{align*}
        &MI(\Delta P^{(S_i)}_j,Y^{(S_i)}_j)\\&=H(Y^{(S_i)}_j)+\sum_{X^{(S_i)}_j,Y^{(S_i)}_j}\sum_{\Delta P^{(S_i)}_j\in \mathbb{V}(X^{(S_i)}_j)}p(\Delta P^{(S_i)}_j,Y^{(S_i)}_j)\\&\log \frac{p(X^{(S_i)}_j,Y^{(S_i)}_j)}{p(X^{(S_i)}_j)}\\&=H(Y^{(S_i)}_j)+\sum_{X^{(S_i)}_j,Y^{(S_i)}_j}[\sum_{\Delta P^{(S_i)}_j\in \mathbb{V}(X^{(S_i)}_j)}p(\Delta P^{(S_i)}_j,Y^{(S_i)}_j)]\\&\log \frac{p(X^{(S_i)}_j,Y^{(S_i)}_j)}{p(X^{(S_i)}_j)}\\&=H(Y^{(S_i)}_j)+\sum_{X^{(S_i)}_j,Y^{(S_i)}_j}p(X^{(S_i)}_j,Y^{(S_i)}_j)\log \frac{p(X^{(S_i)}_j,Y^{(S_i)}_j)}{p(X^{(S_i)}_j)}\\&=MI(X^{(S_i)}_j,Y^{(S_i)}_j).
    \end{align*}
\end{proof}
\begin{property}[Adding New Information] By minimizing Equation \eqref{eq:fidelity loss}, the generated prompt $\Delta P^{(S_i)}_j$ contains new information comparing to the time series input $X^{(S_i)}_j$, \textit{i.e.}, $H(\Delta P^{(S_i)}_j)\geq H(X^{(S_i)}_j)$.
\end{property}
\begin{proof}
    Without loss of generality, we assume that a finite number of prompts are generated for each time series input, and each prompt is generated independently. Then we have $p(X^{(S_i)}_j)=\sum_{\Delta P^{(S_i)}_j\in \mathbb{V}(X^{(S_i)}_j)} p(\Delta P^{(S_i)}_j)$. It follows that
    \begin{align*}
        &H(X^{(S_i)}_j)\\&=-\sum_{X^{(S_i)}_j}p(X^{(S_i)}_j)\log(p(X^{(S_i)}_j))\\&=-\sum_{X^{(S_i)}_j}[\sum_{\Delta P^{(S_i)}_j\in \mathbb{V}(X^{(S_i)}_j)} p(\Delta P^{(S_i)}_j)]\log([\sum_{\Delta P^{(S_i)}_j\in \mathbb{V}(X^{(S_i)}_j)} p(\Delta P^{(S_i)}_j)])\\&=-\sum_{X^{(S_i)}_j}\sum_{\Delta P^{(S_i)}_j\in \mathbb{V}(X^{(S_i)}_j)} p(\Delta P^{(S_i)}_j)\log([\sum_{\Delta P^{(S_i)}_j\in \mathbb{V}(X^{(S_i)}_j)} p(\Delta P^{(S_i)}_j)])\\&\leq - \sum_{X^{(S_i)}_j}\sum_{\Delta P^{(S_i)}_j\in \mathbb{V}(X^{(S_i)}_j)} p(\Delta P^{(S_i)}_j)\log( p(\Delta P^{(S_i)}_j))\quad \text{(Jensen's Inequality)}\\&=-\sum_{\Delta P^{(S_i)}_j\in \mathbb{V}(X^{(S_i)}_j)} p(\Delta P^{(S_i)}_j)\log( p(\Delta P^{(S_i)}_j))=H(\Delta P^{(S_i)}_j)
    \end{align*}
\end{proof}
\subsection{Proofs of Theorems \ref{theo: universality} and \ref{theo: flexibility}}
\label{sec:theorem proof}
 To prove Theorems \ref{theo: universality} and \ref{theo: flexibility}, we follow the similar procedure of \cite{wang2023universality}. To make proofs self-contained, we first mathematically formulate our simplified POND model $f$. Without loss of generality, we assume that $f$ has only one expert transformer network, and it consists of an attention layer and an MLP layer. The attention layer and the transformer layer are defined as follows \cite{wang2023universality}:
\begin{definition} [Attention Layer]
    The $\textbf{h}$-head attention layer between a time-stamp $\textbf{x}$ and a time series $\textbf{X}$ is defined as follows:
    \begin{align*}
Att(\textbf{x},\textbf{X})=\sum_{i=1}^\textbf{h} \textbf{W}^i_o\textbf{W}^i_v\textbf{X}\sigma((\textbf{W}^i_k\textbf{X})^T\textbf{W}^i_q\textbf{x})
    \end{align*}
where $\textbf{W}^i_q$, $\textbf{W}^i_k$, $\textbf{W}^i_v$ and $\textbf{W}^i_o(i=1,\cdots,\textbf{h})$ are parameterized weights, and $\sigma$ is a softmax operator. The normalizing factor of $\frac{1}{\sqrt{\textbf{d}_{kq}}}$ is subsumed in the weight matrices $\textbf{W}^i_k$ for notational simplicity.
\end{definition}
 We then define the cross-attention between two time series $\textbf{X}\in \mathbb{R}^{n\times L}$ and $\textbf{X}^{'}\in \mathbb{R}^{n\times L}$:
\begin{align*}
    Att(\textbf{X},\textbf{X}^{'})=[Att(\textbf{X}_{:,1},\textbf{X}^{'}),Att(\textbf{X}_{:,2},\textbf{X}^{'}),\cdots,Att(\textbf{X}_{:,L},\textbf{X}^{'})]
\end{align*}
where $\textbf{W}_{:,j}$ is the $j$-th column of $\textbf{W}$.
\begin{definition}[Simplified POND Model]
    The simplified POND model $f$ is shown as follows:
    \begin{align*}
        MLP(\textbf{X})&=[\textbf{W}_2 RELU(\textbf{W}_1\textbf{X}_{:,1})+\textbf{b}_1+\textbf{b}_2+\textbf{X}_{:,1},\cdots,\\&\textbf{W}_2 RELU(\textbf{W}_1\textbf{X}_{:,n})+\textbf{b}_1+\textbf{b}_2+\textbf{X}_{:,n}]\\
        f(\textbf{X})&=MLP(Att(\textbf{X},\textbf{X})+\textbf{X}).
    \end{align*}
    where $RELU(\cdot)$ is the ReLU activation function.
\end{definition}
\begin{theorem}[Universality of our POND Model]
    Let $1\leq q<\infty$ and $\varepsilon>0$, and $\mathcal{F}^{(S_i)}: [0,1]^{n\times L}\rightarrow [0,1]^{\vert C\vert}$ is a time series classifer, which is trained from source domain $S_i$ and is $\mathcal{L}$-Lipschitz, there exist a prompt length $m$ and a POND model $f$ such that for any  $\mathcal{F}^{(S_i)}$,  we can find a domain-specific prompt generator $g^{(S_i)}: [0,1]^{n\times L}\rightarrow \mathbb{R}^{n\times m}$ from source domain $S_i$ with $d_q(f([P+g^{(S_i)}(\cdot),\cdot]),\mathcal{F}^{(S_i)})<\varepsilon$ for all $S_i(i=1,2,\cdots M)$.
\end{theorem}
\begin{proof}
    Let the common prompt $P=0$, the prompt generator $g^{(S_i)}$ be constant $P^{(S_i)}$, and $f$ be a transformer with two heads of size one and four hidden units, then this theorem can be directly derived from Theorem 1 in \cite{wang2023universality}.
\end{proof}
To prove Theorem \ref{theo: flexibility}, we need two assumptions on our simplified POND model $f$, which are shown as follows:
\begin{assumption}[Assumption on the Attention Layer] $Att(X^{(S_1)}_1,X^{(S_1)}_1)+X^{(S_1)}_1\neq Att(X^{(S_2)}_1,X^{(S_2)}_1)+X^{(S_2)}_1$ in Theorem \ref{theo: flexibility}, and
$\textbf{W}_o$, $\textbf{W}_k$, $\textbf{W}_q$, and $\textbf{W}_v$ are full rank.  \label{ass:attention}
\end{assumption}
\begin{assumption}[Assumption on the MLP Layer]
        $Y^{(S_i)}_1(i=1,2)$ in Theorem \ref{theo: flexibility} are in the range set of $MLP$. Moreover, the number of channels $n\geq 2+dim((MLP^{-1}(Y^{(S_1)}_1)-\mathcal{X}_1)\cup (MLP^{-1}(Y^{(S_2)}_1)-\mathcal{X}_2))$ in Theorem \ref{theo: flexibility}. Here $dim(\textbf{S})$ measures the dimension of the subspace spanned by vectors in a set $\textbf{S}$ and $MLP^{-1}(\textbf{y})=\{\textbf{x}: MLP(\textbf{x})=\textbf{y}\}$. \label{ass:MLP}
\end{assumption}
Aside from two assumptions, the following Lemma is also useful to prove Theorem \ref{theo: flexibility}.
\begin{lemma}(Lemma 7 in \cite{wang2023universality})
    Given $\textbf{c}\in\mathbb{R}^{n\times L}$ and full-rank attention weights $\textbf{W}_q$, $\textbf{W}_k$, and $\textbf{W}_v$, there are $\textbf{x}_0$ almost everywhere for which  there exists $\textbf{x}_1\in\mathbb{R}^{n\times L}$ such that $Att(\textbf{x}_0,[\textbf{x}_0,\textbf{x}_1])||\textbf{c}$.
\end{lemma}
\begin{theorem}[Flexibility of of our POND Model]
    Consider two labeled time series pairs $(X^{(S_1)}_1=[\mathcal{X}_1,\mathcal{X}_0],Y^{(S_1)}_1)$ and $(X^{(S_2)}_1=[\mathcal{X}_2,\mathcal{X}_0],Y^{(S_2)}_1)$  from two source domains $S_1$ and $S_2$, respectively, where $Y^{(S_1)}_1\neq Y^{(S_1)}_2$. For some proposed POND model $f$:\\
    (a).$[$The limitation of prompt tuning$]$There exists no prompt $P$ such that $f([P,X^{(S_i)}_1])=Y^{(S_i)}_1(i=1,2)$.\\
    (b).$[$Our POND Model handles this limitation$]$ There exist the common prompt $P$ and the prompt generators $g^{(S_i)}(i=1,2)$ such that \\$f([P+g^{(S_i)}(X^{(S_i)}_1),X^{(S_i)}_1])=Y^{(S_i)}_1(i=1,2)$.
\end{theorem}
\begin{proof}
(a). Firstly, we consider the prompt $P$ only (\textit{i.e.}, without  the prompt generator $g^{(S_i)}$), we pass $X^{(S_1)}_1$ and $X^{(S_2)}_1$ to the attention layer to obtain:
\begin{align}
    \nonumber Att(\mathcal{X}_0,[P,X^{(S_1)}_1])&=\lambda(X^{(S_1)}_1,\mathcal{X}_0,[P,X^{(S_1)}_1])Att(\mathcal{X}_0,X^{(S_1)}_1)\\&+\lambda(P,\mathcal{X}_0,[P,X^{(S_1)}_1])Att(\mathcal{X}_0,P) \label{eq: cone equality 1}\\
    \nonumber Att(\mathcal{X}_0,[P,X^{(S_2)}_1])&=\lambda(X^{(S_2)}_1,\mathcal{X}_0,[P,X^{(S_2)}_1])Att(\mathcal{X}_0,X^{(S_2)}_1)\\&+\lambda(P,\mathcal{X}_0,[P,X^{(S_2)}_1])Att(\mathcal{X}_0,P) \label{eq: cone equality 2}
\end{align}
where $\lambda(\textbf{X},\textbf{X}^{'},\textbf{X}^{''}=[\textbf{X}_1,\textbf{X}_2])\in (0,1)$ is a positive scalar defined as:
\begin{align*}
    \lambda(\textbf{X}_1,\textbf{X}^{'},\textbf{X}^{''})=\frac{\sum_{j}\exp{((\textbf{W}_k\textbf{X}_{:,j}})^T(\textbf{W}_q\textbf{X}^{'}))}{\sum_{j}\exp{((\textbf{W}_k\textbf{X}^{''}_{:,j}})^T(\textbf{W}_q\textbf{X}^{'}))}
\end{align*}
Based on Equations \eqref{eq: cone equality 1} and \eqref{eq: cone equality 2}, we learn that $Att(\mathcal{X}_0,P)$ is the intersection of $Cone(Att(\mathcal{X}_0,[P,X^{(S_1)}_1]),Att(\mathcal{X}_0,X^{(S_1)}_1))$ and\\ $Cone(Att(\mathcal{X}_0,[P,X^{(S_2)}_1],Att(\mathcal{X}_0,X^{(S_2)}_1))$, where $Cone(\textbf{a}_1,\cdot,\textbf{a}_k)=\{x|x=\sum_{i=1}^k c_i\textbf{a}_i, c_i>0(i=1,\cdots,k)\}$ is a convex cone formed by $\textbf{a}_i(i=1,\cdots,k)$. However, due to the same deduction by the proof of Theorem 2 in \cite{wang2023universality}, $Cone(Att(\mathcal{X}_0,[P,X^{(S_1)}_1]),Att(\mathcal{X}_0,X^{(S_1)}_1))$ and $Cone(Att(\mathcal{X}_0,[P,X^{(S_2)}_1],Att(\mathcal{X}_0,X^{(S_2)}_1))$ have no intersection based on Assumption \ref{ass:MLP}, which contradicts the existence of $Att(\mathcal{X}_0,P)$. Therefore, there exists no common prompt $P$ such that $f([P,X^{(S_i)}_1])=Y^{(S_i)}_1(i=1,2)$.\\
(b). Secondly, we illustrate the case when both the common prompt $P$ and  prompt generators $g^{(S_i)}$ are available. In this case, Equations \eqref{eq: cone equality 1} and \eqref{eq: cone equality 2} become the following:
\begin{align}
    &\nonumber Att(\mathcal{X}_0,[P+g^{(S_1)}(X^{(S_1)}_1),X^{(S_1)}_1])\\&\nonumber =\lambda(X^{(S_1)}_1,\mathcal{X}_0,[P+g^{(S_1)}(X^{(S_1)}_1),X^{(S_1)}_1])Att(\mathcal{X}_0,X^{(S_1)}_1)\\&\nonumber+\lambda(P+g^{(S_1)}(X^{(S_1)}_1),\mathcal{X}_0,[P+g^{(S_1)}(X^{(S_1)}_1),X^{(S_1)}_1])\\&Att(\mathcal{X}_0,P+g^{(S_1)}(X^{(S_1)}_1)) \label{eq: domain-specific prompt cone equality 1}\\
    &\nonumber Att(\mathcal{X}_0,[P+g^{(S_2)}(X^{(S_2)}_1),X^{(S_2)}_1])\\& \nonumber=\lambda(X^{(S_2)}_1,\mathcal{X}_0,[P+g^{(S_2)}(X^{(S_2)}_1),X^{(S_2)}_1])Att(\mathcal{X}_0,X^{(S_2)}_1)\\& \nonumber+\lambda(P+g^{(S_2)}(X^{(S_2)}_1),\mathcal{X}_0,[P+g^{(S_2)}(X^{(S_2)}_1),X^{(S_2)}_1])\\&Att(\mathcal{X}_0,P+g^{(S_2)}(X^{(S_2)}_1)) \label{eq: domain-specific prompt cone equality 2}    
\end{align}
Obviously, the role of $g^{(S_i)}$ is to find  $Att(\mathcal{X}_0,P+g^{(S_1)}(X^{(S_1)}_1))\in Cone(Att(\mathcal{X}_0,[P+g^{(S_1)}(X^{(S_1)}_1),X^{(S_1)}_1]),Att(\mathcal{X}_0,X^{(S_1)}_1))$ and $Att(\mathcal{X}_0,P+g^{(S_2)}(X^{(S_2)}_1))\in Cone(Att(\mathcal{X}_0,[P+g^{(S_2)}(X^{(S_2)}_1),X^{(S_2)}_1]),Att(\mathcal{X}_0,X^{(S_2)}_1))$ so that these two cones have no intersections, and therefore the contradiction mentioned in (a) can be addressed. 
\end{proof}

\end{document}